\documentclass[10pt,twocolumn,letterpaper]{article}

\usepackage{cvpr}              

\usepackage[dvipsnames]{xcolor}
\usepackage{xcolor, soul}

\newcommand\method{DiffusionAct}

\definecolor{cvprblue}{rgb}{0.21,0.49,0.74}
\usepackage[pagebackref,breaklinks,colorlinks,citecolor=cvprblue]{hyperref}

\newcommand\blfootnote[1]{%
  \begingroup
  \renewcommand\thefootnote{}\footnote{#1}%
  \addtocounter{footnote}{-1}%
  \endgroup
}

\usepackage[T1]{fontenc}
\usepackage{multirow}
\usepackage{array}
\def\paperID{000} 
\def\confName{CVPR}
\def\confYear{2024}

\title{\method: Controllable Diffusion Autoencoder for One-shot Face Reenactment}
\author{\parbox{16cm}{\centering
    {\large Stella Bounareli$^1$,  Christos Tzelepis$^2$,  Vasileios Argyriou$^1$,  Ioannis Patras$^3$, \\ Georgios Tzimiropoulos$^3$ }\\
    {\normalsize
    $^1$ School of Computer Science and Mathematics, Kingston University London\\
    $^2$ School of Science and Technology, City University of London\\
    $^3$ School of Electronic Engineering and Computer Science, Queen Mary University of London }}
}

\begin{document}

\twocolumn[{%
\renewcommand\twocolumn[1][]{#1}%
\vspace{-1em}
\maketitle
    \begin{center}
    \centering 
    \includegraphics[width=1.0\linewidth]{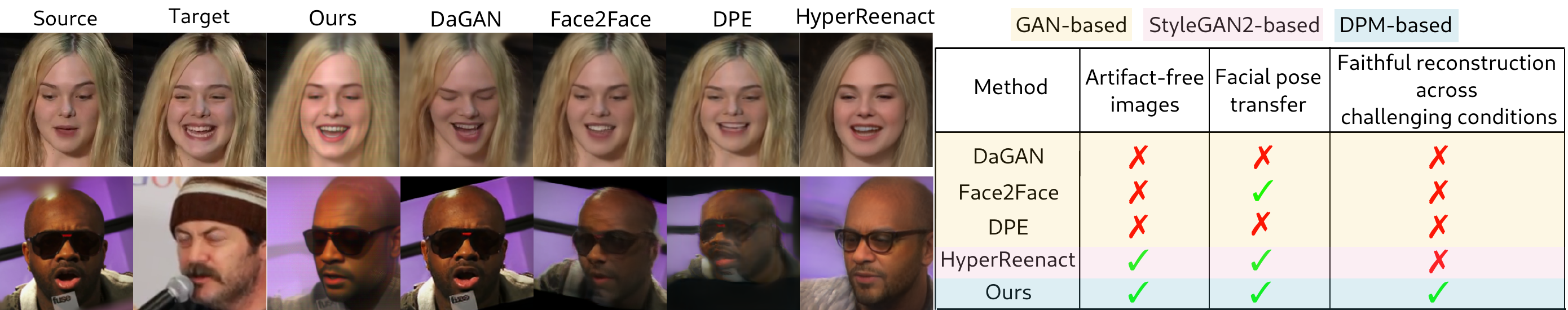}
    \captionof{figure}{Our DPM-based method, \method, performs \textit{one-shot} self (top row) and cross-subject (bottom row) neural face reenactment. We demonstrate that, compared to current state-of-the-art methods, namely DaGAN~\cite{hong2022depth}, Face2Face~\cite{yang2022face2face}, DPE~\cite{pang2023dpe} and HyperReenact~\cite{bounareli2023hyperreenact}, \method~produces realistic, artifact-free images, accurately transfers the target head pose and expression, and faithfully reconstructs the source identity and appearance across challenging conditions, e.g., large head pose movements.}
    \label{fig:intro}
    \end{center}
}]
\maketitle

\begin{abstract}
Video-driven neural face reenactment aims to synthesize realistic facial images that successfully preserve the identity and appearance of a source face, while transferring the target head pose and facial expressions. Existing GAN-based methods suffer from either distortions and visual artifacts or poor reconstruction quality, i.e., the background and several important appearance details, such as hair style/color, glasses and accessories, are not faithfully reconstructed. Recent advances in Diffusion Probabilistic Models (DPMs) enable the generation of high-quality realistic images. To this end, in this paper we present \method, a novel method that leverages the photo-realistic image generation of diffusion models to perform neural face reenactment. Specifically, we propose to control the semantic space of a Diffusion Autoencoder (DiffAE), in order to edit the facial pose of the input images, defined as the head pose orientation and the facial expressions. Our method allows one-shot, self, and cross-subject reenactment, without requiring subject-specific fine-tuning. We compare against state-of-the-art GAN-, StyleGAN2-, and diffusion-based methods, showing better or on-par reenactment performance. \blfootnote{Project page: \url{https://stelabou.github.io/diffusionact/}}
\end{abstract}

\section{Introduction}
\label{sec:intro}

    Neural face reenactment aims to generate photo-realistic head avatars of a source subject. That is, given a source facial image, the goal of face reenactment is to animate it using either audio driving signals~\cite{prajwal2020lip, gan2023efficient, guan2023stylesync,shen2023difftalk, stypulkowski2023diffused} or video driving sequences~\cite{zakharov2020fast, wang2021one,meshry2021learned,bounareli2023hyperreenact} from the same (self reenactment), or different subjects (cross-subject reenactment), as shown in Fig.~\ref{fig:intro}. In doing so, the appearance and the identity characteristics of the source face should be preserved, while the target facial pose (which in the literature of neural face reenactment is typically defined as the head pose orientation and the facial expressions~\cite{burkov2020neural,zakharov2020fast,bounareli2023hyperreenact}) is accurately transferred to the reenacted face. Neural face reenactment has recently drawn significant attention both from the research community and the industry, due to its wide range of applications, e.g. in entertainment, creative arts, video conferencing and video production.  

    The advent of Generative Adversarial Networks (GANs)~\cite{goodfellow2014generative,karras2017progressive,karras2019style} and Diffusion Probabilistic Models (DPMs)~\cite{ho2020denoising, Song2021ddim, rombach2022high} has led to impressive performance in the tasks of image synthesis and manipulation. Such methods are able to produce realistic and aesthetically pleasing synthetic images, but also provide controllability over the generative process~\cite{voynov2020unsupervised,tzelepis2021warpedganspace,oldfield2023panda,tzelepis2022contraclip,patashnik2021styleclip,huang2023collaborative,chen2023face}. Whilst recent generative methods have been extensively used for face reenactment~\cite{zakharov2020fast, burkov2020neural, siarohin2019first}, it remains a remarkably challenging task due to the wide range of variations of the human faces, different facial shapes, appearances and expressions, head pose movements and lighting conditions. A certain line of research proposes training controllable GAN models~\cite{zakharov2020fast,doukas2020headgan,meshry2021learned,ren2021pirenderer}. However, training such models often leads to deformations and visual artifacts on the generated images. As shown in Fig.~\ref{fig:intro}, GAN-based methods~\cite{hong2022depth, yang2022face2face, pang2023dpe} often fail to produce realistic images under head pose movements between the source and target faces. Furthermore, learning disentangled representations of identity/appearance and facial pose, which is of paramount importance for face reenactment, is a non trivial task. Failure in achieving disentanglement can lead to identity leakage from the driving faces to the source ones~\cite{zakharov2020fast,zakharov2019few}. To overcome the above challenges, recent methods~\cite{bounareli2022finding,yin2022styleheat,bounareli2023hyperreenact,oorloff2023robust} focus on leveraging the photo-realistic generative ability and the disentangled properties of pre-trained StyleGAN2 models~\cite{karras2020analyzing}. Nevertheless, a significant limitation of these approaches is that the encoding of real images onto StyleGAN2 space remains a challenge, leading usually to unfaithful reconstruction and poor reenactment. As illustrated in Fig.~\ref{fig:intro}, the StyleGAN2-based method of HyperReenact~\cite{bounareli2023hyperreenact} fails to precisely reconstruct appearance details, such as hair styles and glasses, that are crucial to maintain the overall perceptual quality of the reenacted images. Recent works incorporate DPMs for the task of audio-driven talking head sequences generation~\cite{shen2023difftalk,stypulkowski2023diffused,du2023dae}. However, these methods mainly focus on editing the mouth region. In contrast, DiffusionRig~\cite{ding2023diffusionrig} enables editing the head pose and expression of input images, however, it requires subject-specific fine-tuning with around 20 images to reconstruct the source identity characteristics, limiting its feasibility. Finally, IP-Adapter~\cite{ye2023ip} and InstantID~\cite{wang2024instantid} condition Stable Diffusion (SD) models~\cite{rombach2022high} to generate controllable images. Nevertheless, such methods are not able to preserve both the identity and appearance of input faces, making them impractical for the task of face reenactment.

    In this work, we propose a novel neural face reenactment framework, called \method, which incorporates the generative power of a pre-trained Denoising Diffusion Implicit Model (DDIM)~\cite{Song2021ddim} and the remarkable reconstruction performance of a Diffusion AutoEncoder (DiffAE)~\cite{preechakul2022diffusionauto}. The latter learns a semantic encoder that maps input images into meaningful semantic codes, which then are decoded by a standard DDIM sampler, allowing for image editing and, thus, can be employed for face reenactment. In contrast to StyleGAN2-based methods~\cite{bounareli2022finding, yin2022styleheat, bounareli2023hyperreenact}, that typically suffer from the reconstruction-editability trade-off~\cite{tov2021designing}, our method achieves effective face reenactment performance relying on the near-perfect reconstruction capability of DiffAE. Furthermore, our method requires a single frame from the source face, in contrast to other GAN- or DPM-based methods~\cite{burkov2020neural,hsu2022dual,ding2023diffusionrig}. Concretely, we devise a conditioning mechanism on the semantic encoder of DiffAE, which we call \textit{reenactment encoder}, that enables the transfer of the target facial pose into the reenacted image. More specifically, we optimise our reenactment encoder to encode in its semantic space the appearance/identity characteristics of the source face, and the target facial pose, given in terms of informative facial landmarks. The resulting ``reenacted'' code is then decoded by DDIM to generate the reenacted face.
   
    The main contributions of this paper can be summarized as follows:
    \begin{enumerate}
        \item We introduce \method, a novel approach for face reenactment based on a pre-trained diffusion autoencoder model~\cite{preechakul2022diffusionauto}. To the best of our knowledge, our work is the first that uses a pre-trained DPM and adapts it to the face reenactment task. 
        
        \item Inspired by ControlNet~\cite{zhang2023controlnet}, we propose a conditioning mechanism on a pre-trained semantic encoder to enable the generation of controllable images using the target facial landmarks. In contrast to ControlNet~\cite{zhang2023controlnet}, that requires the costly optimisation of the DPM's UNet~\cite{ronneberger2015u}, our method optimises only the reenactment encoder.
        
        \item We report extensive quantitative and qualitative results on VoxCeleb1~\cite{Nagrani17} and VoxCeleb2~\cite{Chung18b} datasets, both on self and cross-subject reenactment and we compare against 5 state-of-the-art GAN-based methods, 3 StyleGAN2-based methods and 1 DPM-based method. We demonstrate that our method effectively tackles the challenging task of neural face reenactment by generating artifact-free images and by accurately transferring the target facial pose and faithfully reconstructing the identity and appearance characteristics of the source faces across different challenging conditions.
    \end{enumerate}

\section{Related work}
\label{sec:related_work}

    \noindent \textbf{Diffusion Probabilistic Models (DPMs)}: DPMs~\cite{ho2020denoising,Song2021ddim, dhariwal2021diffusion} are generative models that learn a data distribution through a denoising process that iteratively transforms noisy samples into the target distribution. The recently introduced Latent Diffusion Models (LDMs)~\cite{rombach2022high}, transfer the denoising process into a lower-dimension latent space, enabling both faster training and better image generation performance. Currently, DPMs excel in conditional image generation~\cite{rombach2022high, zhang2023controlnet,ponglertnapakorn2023difareli, ye2023ip, wang2024instantid} and image editing~\cite{lugmayr2022repaint, sedit2022, kim2022diffusionclip, mokady2023null,pan2023effective,huang2023collaborative, kim2023diffusion}. A recent approach for conditioning a DPM on various modalities (such as segmentation maps and keypoints) is ControlNet~\cite{zhang2023controlnet}, that does so by learning to adapt a pre-trained diffusion model on the aforementioned additional conditions. Finally, Preechakul et al.~\cite{preechakul2022diffusionauto} proposed DiffAE, a method that learns an image encoder to predict a semantic code which is then used to condition the diffusion process. Similarly to StyleGAN2-based methods~\cite{voynov2020unsupervised, 2020ganspace}, DiffAE allows for semantic/latent code editing which leads to meaningful changes in the input images~\cite{preechakul2022diffusionauto, yue2023chatface}, while in contrast to SD, DiffAE enables accurate image reconstruction. 
   
    \begin{figure*}[t]
        \centering
        \includegraphics[width=1.0\textwidth]{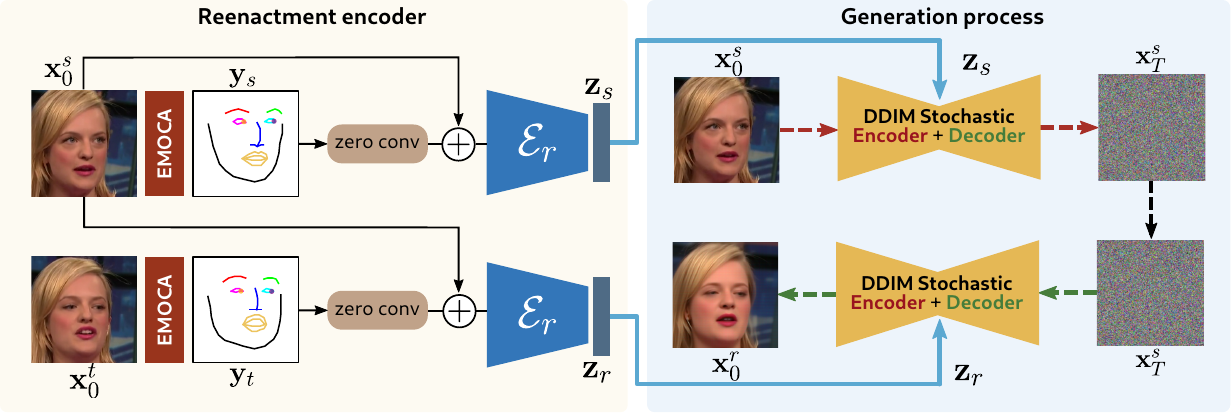}
        \caption{\textbf{Overview of the proposed method:} Given a pair of a source ($\mathbf{x}_0^s$) and a target ($\mathbf{x}_0^t$) images, we condition the reenactment encoder $\mathcal{E}_r$ on the target facial landmarks $\mathbf{y}_t$, in order to predict the reenactment semantic code $\mathbf{z}_r$ that, when decoded by the DDIM, generates the reenacted image $\mathbf{x}_0^r$ that captures the source identity/appearance and the target head pose and facial expressions.}
        \label{fig:architecture}
    \end{figure*}

    \noindent \textbf{Neural face reenactment}: The majority of existing face reenactment methods train controllable GANs conditioned on various facial information or interest (such as facial landmarks or keypoints). There is a line of works that propose to drive the image generation process using facial landmarks~\cite{zakharov2020fast,zakharov2019few}, extracted from off-the-shelf facial landmark estimation methods such as~\cite{bulat2017far}. Nevertheless, these methods suffer from identity leakage from the target faces to the reenacted ones on cross-subject reenactment, as facial landmarks typically maintain the facial shape of the input faces. Several approaches~\cite{doukas2020headgan,bounareli2022finding,bounareli2023hyperreenact,ren2021pirenderer, khakhulin2022rome} leverage the disentangled properties of 3D Morphable Models (3DMMs)~\cite{blanz1999morphable} to drive the generation process. Utilizing 3DMMs allows for disentanglement of the facial shape from the facial pose, which mitigates the identity leakage on cross-subject reenactment. Additionally, warping-based methods rely on learning keypoints that represent the facial motion between the source and target faces~\cite{siarohin2019first,wang2021one, zhao2022thin, hong2022depth}. Specifically, FOMM~\cite{siarohin2019first} proposes to model the motion change between the source and the target faces by calculating the changes in the position of the learned keypoints. This method can only work on sequences with minor changes of head pose, generating many visual artifacts and deformations otherwise. Several methods build upon FOMM~\cite{siarohin2019first} by introducing 3D keypoints~\cite{wang2021one} or using depth-guided facial keypoints~\cite{hong2022depth}. Nevertheless, the warping-based approaches still produce unnatural deformations on the reenacted images.
    
    A recent line of works~\cite{bounareli2023stylemask,bounareli2022finding,bounareli2023hyperreenact,yin2022styleheat,oorloff2023robust} propose to leverage the photo-realistic image generation of pre-trained StyleGAN2 models~\cite{karras2020analyzing}. Bounareli et al.~\cite{bounareli2022finding} investigate the $\mathcal{W}$ latent space of StyleGAN2 to find the directions that are responsible for facial pose editing. In a similar fashion, Oorloff et al.~\cite{oorloff2023robust} propose to use a combination of $\mathcal{W}$ space and  style space $\mathcal{S}$ of StyleGAN2. Such methods are able to faithfully edit real images, however they require to optimize the weights of the GAN generator~\cite{roich2021pivotal} to preserve the appearance of the source images. StyleHeat~\cite{yin2022styleheat} proposes to use the feature space $\mathcal{F}$ of StyleGAN2, which despite improving reconstruction, leads to poor reenactment performance even under moderate head pose variations. Finally, HyperReenact~\cite{bounareli2023hyperreenact}, optimizes a hypernetwork~\cite{hypernets} to alter the weights of StyleGAN2 generator, simultaneously refining the source appearance and transferring the target facial pose. In spite of the ability of HyperReenact to faithfully transfer the target head pose and expression, it often tends to produce over-smoothed reenacted images, missing important details of the source image (e.g., glasses, background or hair styles).
    
    Finally, recent works train conditional DPMs to generate \textit{audio-driven} talking head sequences~\cite{shen2023difftalk,stypulkowski2023diffused,du2023dae}. However, audio-driven methods typically focus on reenacting the mouth region only, limiting their ability to edit the head pose and facial expressions. DiffusionRig~\cite{ding2023diffusionrig} enables editing both the head pose and facial expressions by learning a DPM driven by a 3DMM~\cite{feng2020deca}. However, it requires subject-specific fine-tuning with a small portrait dataset to accurately reconstruct the source appearance and identity. Additionally, the authors of~\cite{xu2023multimodal} learn to animate a source face using multimodal guidance from audio signals and 3D parameters. Nevertheless, all the aforementioned methods require training a diffusion model which is computationally expensive. On the contrary, in this work we propose to use a pre-trained DPM and condition the generation process in order to perform one-shot face reenactment.

\section{Proposed method}
\label{sec:method}

    In this section we present \method~for one-shot neural face reenactment. An overview of the proposed method is illustrated in Fig.~\ref{fig:architecture}. Our method builds on a pre-trained DiffAE~\cite{preechakul2022diffusionauto}, where we propose to learn a ``reenactment encoder'' $\mathcal{E}_r$ and condition it on the target facial landmarks $\mathbf{y}_t$ so as to learn to predict the semantic code $\mathbf{z}_r$ that reenacts, via the generative DDIM module, the source image $\mathbf{x}_0^s$ given a target image $\mathbf{x}_0^t$. This way, the reenacted image $\mathbf{x}_0^r$ preserves the identity/appearance of the source image, while it transfers the head orientation and facial expressions of the target image. In the rest of the section, we first briefly present the adopted DiffAE method (Sect.~\ref{ssec:background}), and then we describe the proposed framework (Sect.~\ref{ssec:method}) and the adopted training protocol (Sect.~\ref{ssec:training_protocol}).

\subsection{Background}
\label{ssec:background}
    
    We briefly discuss DiffAE, a recent method that equips a standard DDIM sampler with the editing property via a semantic vector space (similar to the latent space of a GAN). DiffAE utilizes a conditional DDIM~\cite{Song2021ddim} that learns to iteratively transform noisy samples into images that resemble a target distribution.

    In contrast to a Denoising Diffusion Probabilistic Model (DDPM)~\cite{ho2020denoising} that uses a Markov Chain model to progressively add noise on a real image $q(\mathbf{x}_t | \mathbf{x}_{t-1})$, DDIM speeds up the generation process by defining it as a non-Markovian process $q(\mathbf{x}_t | \mathbf{x}_{t-1}, \mathbf{x}_0)$, further conditioned on the input image $\mathbf{x}_0$. The reverse (denoising) process $p(\mathbf{x}_{t-1} | \mathbf{x}_t)$ is learned using a function $\epsilon_{\theta}(\mathbf{x}_t, t)$ that takes as input a noisy image $\mathbf{x}_t$ and learns to predict the added noise $\epsilon$ to $\mathbf{x}_0$ in order to get $\mathbf{x}_t$. This function is modeled as a UNet~\cite{ronneberger2015u} and is trained via minimising $\lVert\epsilon_{\theta}(\mathbf{x}_t, t) - \epsilon \rVert$. The denoised image at step $t-1$ is calculated as
    \begin{equation}
        \mathbf{x}_{t-1} = \sqrt{a_{t-1}} f_{\theta}(\mathbf{x}_t, t) + \sqrt{1-a_{t-1}} \epsilon_{\theta}(\mathbf{x}_t, t),
    \end{equation}\label{eq:denoise_process}
    where $f_{\theta}(\mathbf{x}_t, t)$ is the estimated real image $\hat{\mathbf{x}}_0$ at step $t$,
    \begin{equation}
       f_{\theta}(\mathbf{x}_t, t) = \frac{\mathbf{x}_t - \sqrt{1-a_t}\epsilon_{\theta}(\mathbf{x}_t, t)}{\sqrt{a_t}},
    \end{equation}
    and $a_t = \prod_{s=1}^{t} (1-\beta_s)$, where $\beta_t$ is a fixed or learned variance schedule that represents the noise levels. The generation process of a DDIM can be deterministic, meaning that there is a deterministic mapping between the real image $\mathbf{x}_0$ and the noisy image $\mathbf{x}_T$, and vice versa.

    For endowing a DDIM with a semantic space, DiffAE learns a semantic encoder $\mathcal{E}$ that learns to encode an image $\mathbf{x}_0$ into a semantic code $\mathbf{z}\in\mathbb{R}^{512}$. The diffusion process $\epsilon_{\theta}(\mathbf{x}_t, t, \mathbf{z})$ is then conditioned on the semantic code $\mathbf{z}$. Using the semantic encoder $\mathcal{E}$ and the DDIM sampler, a real image $\mathbf{x}_0$ is mapped onto the stochastic subcode $\mathbf{x}_T$ using the reverse process of (\ref{eq:denoise_process}):
    \begin{equation}
       \mathbf{x}_{t+1} = \sqrt{a_{t+1}} f_{\theta}(\mathbf{x}_t, t, \mathbf{z}) + \sqrt{1-a_{t+1}} \epsilon_{\theta}(\mathbf{x}_t, t, \mathbf{z}).
    \end{equation}\label{eq:noise_process}
    Finally, a real image is reconstructed remarkably faithfully by the DDIM of DiffAE via decoding its semantic code $\mathbf{z}$ and stochastic subcode $\mathbf{x}_T$.

\subsection{\method}
\label{ssec:method}

    In this section we present our framework for neural face reenactment. The proposed method employs a DDIM (pre-trained on the FFHQ~\cite{karras2019style} dataset) and a semantic encoder, which --in contrast to DiffAE that aims to predict a semantic code encoding the input image-- learns to predict a semantic code that encodes the pose of a target face, and the appearance and identity characteristics of a source face. For doing so, we propose to condition the semantic encoder, which hereinafter will be referred to as the \text{reenactment encoder}, on the facial landmarks and the gaze direction of the target face, as shown in Fig.~\ref{fig:architecture}.
    
    Concretely, given a pair of source ($\mathbf{x}_0^s$) and target ($\mathbf{x}_0^t$) images, we aim to generate the reenacted $\mathbf{x}_0^r$ image, that preserves the source appearance and identity characteristics (i.e., background, hair style/color, facial accessories) and transfers the target facial pose (i.e., head pose orientation and facial expressions). To do so, we propose learning to predict the desired reenacted semantic code $\mathbf{z}_r$ by conditioning the reenactment encoder $\mathcal{E}_r$ on the facial landmarks and the gaze position of the target face, $\mathbf{y}_t$, extracted by the pre-trained EMOCA~\cite{danvevcek2022emoca} network and the pre-trained gaze estimation network of~\cite{zhang2020eth}. We detail the construction of $\mathbf{y}_t$ in the supplementary material.

    For conditioning the reenactment encoder ($\mathcal{E}_r$) on the target facial pose ($\mathbf{y}_t$), inspired by the recently published ControlNet~\cite{zhang2023controlnet}, we introduce the target landmark condition through a ``zero'' convolution, the output of which we add to the source image before we feed it to the reenactment encoder $\mathcal{E}_r$, as shown in left part of Fig.~\ref{fig:architecture}. We note that, according to~\cite{zhang2023controlnet}, a ``zero'' convolution refers to an $1\times1$ convolution layer whose weights are only initially set to zero and allowed to gradually change during training injecting the desired spatial landmark-based condition into the training of the proposed reenactment encoder $\mathcal{E}_r$. Finally, as illustrated in the right part of Fig.~\ref{fig:architecture}, in order to generate the reenacted image, we first encode the source image $\mathbf{x}_0^s$ into the stochastic noisy subcode $\mathbf{x}_T^s$ using Eq.~\ref{eq:noise_process} and the source semantic code $\mathbf{z}_s$. Given the noisy subcode $\mathbf{x}_T^s$ and the reenacted code $\mathbf{z}_r$, the DDIM decoder generates the reenacted image $\mathbf{x}_0^r$.

\subsection{Training protocol}
\label{ssec:training_protocol}

    In this section we present the training process of the proposed framework which is split into two stages: i) first, we pre-train the reenactment encoder $\mathcal{E}_r$ under the self reenactment setting, and ii) we continue training the reenactment encoder $\mathcal{E}_r$ by incorporating the DDIM sampler to synthesize the reenacted images.

    \textbf{Pre-training:} We first pre-train the reenactment encoder $\mathcal{E}_r$ to predict the reenacted semantic code $\mathbf{z}_r$ that encodes the identity/appearance of the source face and the facial pose of the target face. We do so under the self-reenactment setting, i.e., where the source and target faces have the same identity --  i.e., we first learn to predict a reenacted code $\mathbf{z}_r$ that is identical to the code of the target face, $\mathbf{z}_t$, minimising their $\ell_1$ distance, $\lVert \mathbf{z}_r - \mathbf{z}_t \rVert_1$. We note that we initialize the reenactment encoder $\mathcal{E}_r$ by adopting the weights of the semantic encoder $\mathcal{E}$ of DiffAE. Additionally, the DDIM sampler is not part of this pre-training process.

    After training the proposed reenactment encoder $\mathcal{E}_r$ using only this objective, we obtain an encoder that predicts a semantic code $\mathbf{z}_r$ encoding both the appearance of the source face and the target facial pose. This means that given the predicted code $\mathbf{z}_r$ and the stochastic subcode of the source image $\mathbf{x}_T^s$, the DDIM decoder generates the reenacted image that depicts the source identity in the target facial pose. Whilst this training stage does not result in optimal face reenactment performance (i.e., the reenacted images are prone to some visual artifacts), we show in Sect.~\ref{ssec:ablation} that it is crucial as a warm-up pre-training stage, before we proceed to the main training stage of our framework discussed below.
    
    \textbf{Training:} After obtaining the proposed reenactment encoder $\mathcal{E}_r$ pre-trained on the task of self-reenactment, as discussed above, we proceed to the main training process of $\mathcal{E}_r$ where we additionally incorporate DDIM as a generative module (to generate the reenacted images), along with a series of well-studied reconstruction and pose transfer losses, following the common practice in literature~\cite{bounareli2022finding,xu2022designing,bounareli2023hyperreenact,pang2023dpe}. We illustrate the training process of our method in Fig.~\ref{fig:architecture}. The total loss we minimise is given as:
    \begin{equation}
        \mathcal{L} = \mathcal{L}_{rec} + \mathcal{L}_{pose},
    \end{equation}
    where $\mathcal{L}_{rec}$ and $\mathcal{L}_{pose}$ are the reconstruction and pose losses, respectively, which we discuss briefly below and in detail in the supplementary material.
    We propose the following reconstruction loss to guide the reenacted images to depict the identity and the appearance of the source faces:
    \begin{equation}
        \mathcal{L}_{rec} = 
            \lambda_{pix} \mathcal{L}_{pix} + 
            \lambda_{per} \mathcal{L}_{per} + 
            \lambda_{id} \mathcal{L}_{id} + 
            \lambda_{bg} \mathcal{L}_{bg} +  
            \lambda_{st} \mathcal{L}_{st},
    \end{equation}
    where $\mathcal{L}_{pix}$, $\mathcal{L}_{per}$, and $\mathcal{L}_{id}$ are respectively the pixel-wise, perceptual, and identity losses, calculated as in~\cite{bounareli2023hyperreenact}. We additionally calculate a background loss $\mathcal{L}_{bg}$ by using the off-the-shelf pre-trained face segmentation network~\cite{yu2021bisenet} to create a mask around the face area and keep only the background and hair areas. Finally, inspired by~\cite{barattin2023attribute}, we calculate the style loss $\mathcal{L}_{st}$ as an $\ell_2$ distance in the ViT space of the pre-trained FaRL~\cite{zheng2022general}. 
    Furthermore, in order to force the effective and realistic transfer of the target facial pose (head orientation and facial expressions) onto the reenacted face, we propose to use the following loss term:
    \begin{equation}
        \mathcal{L}_{pose} = 
            \lambda_{g} \mathcal{L}_{g} + 
            \lambda_{sh} \mathcal{L}_{sh} + 
            \lambda_{hp} \mathcal{L}_{hp},
    \end{equation}
    where $\mathcal{L}_{g}$, $\mathcal{L}_{sh}$, and $\mathcal{L}_{hp}$ denote the gaze, shape, and head pose losses, respectively. We calculate the gaze loss as the $\ell_1$ distance of the gaze direction extracted using the off-the-shelf gaze estimation network of~\cite{zhang2020eth}, the shape loss as the $\ell_1$ distance of the 2D facial landmarks~\cite{danvevcek2022emoca}, and the head pose loss as the average $\ell_1$ distance of the three Euler angles (i.e., yaw, pitch, and roll) calculated using the head pose coefficients extracted by EMOCA~\cite{danvevcek2022emoca}. 
        
    Towards achieving improved reenactment performance, we propose a training protocol that involves both self reenactment and image reconstruction tasks. Specifically, during training, we split the mini-batch into two halves. In the first half, the source and target images depict the same subject but on different facial pose (self reenactment), while in the second half, the source and target images are the same (image reconstruction). In our ablation studies (Sect.~\ref{ssec:ablation}), we show that this training protocol improves our qualitative and quantitative results.
        
    Finally, to further improve the performance of the proposed framework, we suggest to fine-tune the DDIM sampler, while being jointly optimized with the proposed reenactment encoder $\mathcal{E}_r$ (see Fig.~\ref{fig:architecture}). We show that this strategy further improves the reenactment performance, since the adopted DDIM~\cite{preechakul2022diffusionauto} has been pre-trained on the FFHQ dataset~\cite{karras2019style}, which has a different distribution of the VoxCeleb datasets~\cite{Nagrani17, Chung18b} in terms of different head poses and expressions~\cite{bounareli2022finding}.

\section{Experiments}
\label{sec:experiments}

    In this section, we provide the implementation and training details of our approach in Sect.~\ref{ssec:details}, we report comparisons of our method with 9 state-of-the-art methods both on self and on cross-subject reenactment tasks in Sect.~\ref{ssec:comparisons}, and we present ablation studies on the impact of the various design choices in Sect.~\ref{ssec:ablation}.

    \subsection{Implementation details}
    \label{ssec:details}
        We use a DDIM model~\cite{preechakul2022diffusionauto} pre-trained on the FFHQ dataset~\cite{karras2019style}. We train our model on the VoxCeleb1 dataset~\cite{Nagrani17}, which contains around 20K videos. During the pre-training stage of the reenactment encoder, we use a learning rate of $10^{-3}$ and a batch size of $32$, while during the main training stage we use $T_{tr}=8$ steps to synthesize the images, a learning rate of $10^{-4}$, and a batch size of $4$. Finally, we fine-tune the DDIM sampler along with the reenactment encoder using a learning rate of $10^{-5}$ and a batch size of $4$. In all training stages we optimize our models using the AdamW~\cite{adamw} optimizer. We also set $ \lambda_{pix}, \lambda_{per}, \lambda_{id}, \lambda_{st} = 20$, $\lambda_{bg} = 10$, $\lambda_{sh} = 0.5$ and $\lambda_{g},\lambda_{hp}=2$. During inference, we use $T=50$ steps to calculate the stochastic subcodes $\mathbf{x}_T^s$ and $T=20$ steps to generate the reenacted images. For our experiments we used 2 Nvidia Quadro RTX 8000.  

    \subsection{Comparisons to state-of-the-art}
    \label{ssec:comparisons}

        In this section we provide quantitative and qualitative comparisons both on self and on cross-subject reenactment against 9 state-of-the-art methods. Specifically, we compare with 5 methods that train controllable GANs from scratch, namely DaGAN~\cite{hong2022depth}, Rome~\cite{khakhulin2022rome}, Face2Face~\cite{yang2022face2face}, UniFace~\cite{xu2022designing} and DPE~\cite{pang2023dpe}. We also provide comparisons with 3 methods that leverage pre-trained StyleGAN2~\cite{karras2020analyzing} models, namely StyleHeat~\cite{yin2022styleheat}, FD~\cite{bounareli2022finding} and HyperReenact~\cite{bounareli2023hyperreenact}, while we also compare with a diffusion-based method called DiffusionRig~\cite{ding2023diffusionrig}. For all methods we use the official publicly available pre-trained models. We evaluated our method on the test set of VoxCeleb1~\cite{Nagrani17}. For each video of the test set we kept the first frame as the source image and the rest as the target images. In the supplementary material we provide additional comparisons on the datasets of VoxCeleb2~\cite{Chung18b} and HDTF~\cite{zhang2021flow} and we also compare against the two SD-based methods namely IP-Adapter~\cite{ye2023ip} and InstantID~\cite{wang2024instantid}.

        \begin{table*}
       
            \begin{center}
            \begin{tabular}{|l|c|c|c|c|c|c|c|c|c|c|c|}
                \hline
                Method & PSNR & SSIM &  CSIM & LPIPS & L1  &  NME &  APD & AED & User Pref. ($\%$) \\
                \noalign{\hrule height 1.2pt}
                DaGAN~\cite{hong2022depth} &  16.9 & 0.73  &\textbf{0.72} & 0.24 & 0.18 & 30.0 & 4.5 & 12.8 & 1.9  \\
                Rome~\cite{khakhulin2022rome} & 7.5 & 0.72 & 0.69 & 0.43 & 0.61 &  13.9 & 1.5 & 8.5  & 7.0  \\
                Face2Face~\cite{yang2022face2face} & 17.1 & 0.76 &  \textbf{0.72} & 0.27 & 0.18 &  15.9 & 1.5 & 10.6 & 10.2 \\
                UniFace~\cite{xu2022designing}  & 18.1 & 0.77 &  0.69 & \textbf{0.22} & 0.17 &  18.5 & 2.3 & 12.7 & 2.0 \\
                DPE~\cite{pang2023dpe} & 14.7 & 0.69 &  0.70 & 0.38 & 0.27 &  23.3 & 4.2 & 11.0 & 5.0 \\
                \hline
                \hline
                StyleHeat~\cite{yin2022styleheat} & 17.5 & 0.79 &   0.58 & 0.28 & 0.18 & 18.8 & 3.4 & 12.0 & 2.9 \\
                FD~\cite{bounareli2022finding} & 18.0 & 0.77 &  0.65 & \underline{0.23} & \underline{0.16} &  14.4 & 1.0 & 9.3 & 4.3  \\
                HyperReenact~\cite{bounareli2023hyperreenact} & \underline{18.9} & \underline{0.80} &  \underline{ 0.71} & 0.25 & \underline{0.16} & \underline{12.6} & \textbf{0.5} & \underline{8.3} & 23.0 \\
                \hline
                \hline
                DiffusionRig~\cite{ding2023diffusionrig} & 15.6 & 0.76 &  0.30 & 0.38 & 0.25 &  18.1 & 1.3 & 11.1 & 0.0\\
                Ours & \textbf{19.7} & \textbf{0.83} &  0.69 & 0.24 & \textbf{0.14}  & \textbf{12.5} & \underline{0.8} & \textbf{7.2} &\textbf{43.7} \\
                \hline
            \end{tabular}
            \end{center}
         \caption{Quantitative results on self reenactment task using the test set of VoxCeleb1 dataset~\cite{Nagrani17}. For CSIM, PSNR and SSIM metrics, higher is better, while for the rest of the metrics lower is better. We show the best and second best results in bold and underline. We note that the User Pref. column corresponds to our user study conducted using both self and cross-subject pairs.}\label{table:metrics_self}
        \end{table*}

\begin{figure*}[h!]
            \centering
            \includegraphics[width=1.0\textwidth]{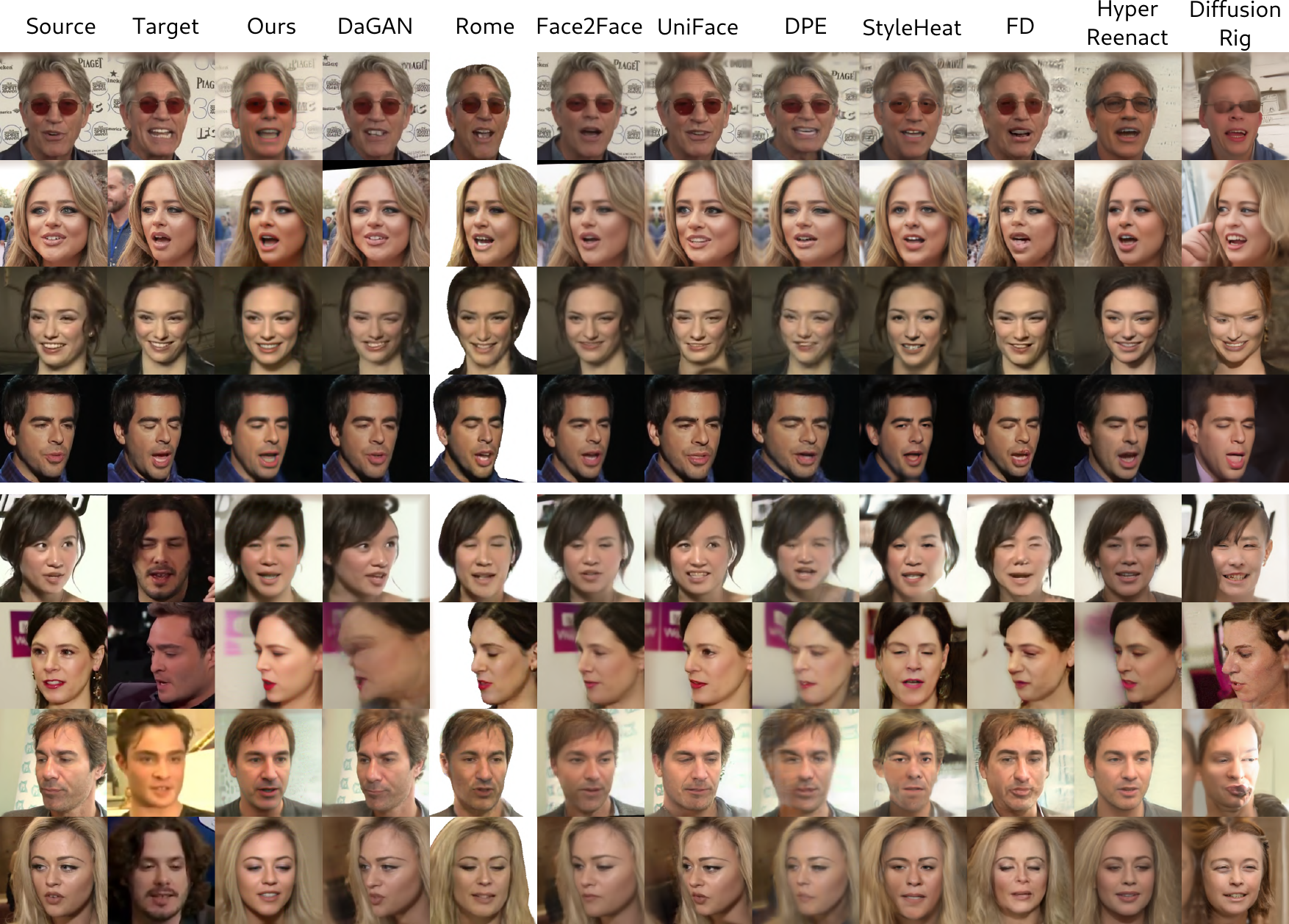}
            \caption{Qualitative comparisons on self (top 4 rows) and cross-subject (bottom 4 rows) reenactment on VoxCeleb1~\cite{Nagrani17} dataset.}
            \label{fig:comparisons}
        \end{figure*}

        \noindent \textbf{Self reenactment task}: On the self reenactment task we report 8 evaluation metrics following the related literature. Specifically, in order to evaluate the reconstruction quality of the presented methods, we calculate the identity similarity (CSIM) using the cosine similarity of the features extracted by ArcFace~\cite{deng2019arcface}, the learned perceptual similarity (LPIPS)~\cite{johnson2016perceptual}, the $\ell_1$ pixel-wise distance (L1), the peak signal-to-noise ratio (PSNR) and the structural similarity index (SSIM). Additionally, to assess the facial pose transfer we calculate the normalized mean error (NME) between the target and the reenacted facial landmarks~\cite{bulat2017far}, the Average Pose Distance (APD) and the Average Expression Distance (AED). For APD and AED, we calculate the average $\ell_1$ distance of the three Euler angles and the expression coefficients extracted using the pre-trained EMOCA~\cite{danvevcek2022emoca} network. In Table~\ref{table:metrics_self}, we present our quantitative results on the self reenactment task using the test set of VoxCeleb1~\cite{Nagrani17}. We note that all metrics are calculated between the target and reenacted images. As shown, our method performs on par with HyperReenact~\cite{bounareli2023hyperreenact} on head pose transfer (APD), while it performs better on expression transfer (AED). Additionally, our method outperforms all other methods on L1, PSNR, and SSIM metrics by generating high quality images that resemble the real ones. Finally, on identity preservation (CSIM) our method remains competitive, however we argue that a higher CSIM score does not necessarily imply better reenactment. Specifically, ArcFace~\cite{deng2019arcface} network, which is used to extract the identity features for calculating the CSIM metric, has been designed and trained to be invariant to certain appearance details (e.g., glasses) or distortions, which are crucial for face reenactment. As shown in the qualitative comparisons (Fig.~\ref{fig:comparisons}), some methods that exhibit higher CSIM scores than our method (such as DaGAN~\cite{hong2022depth} and Face2Face~\cite{yang2022face2face}) tend to produce visual artifacts and deformations, which CSIM fails to capture, while also HyperReenact~\cite{bounareli2023hyperreenact} fails to accurately reconstruct the source appearance details, e.g., glasses, hair styles (Fig.~\ref{fig:comp_hyperreenact}). In the supplementary material we provide visual comparisons against the methods that have higher CSIM scores. As shown, while our reenacted images are visually superior, the CSIM score is lower compared to the other methods.

        \noindent \textbf{Cross-subject reenactment task} On the cross-subject reenactment task, the source and target faces are from different subjects, as a result the evaluation metrics that we report here are the identity similarity (CSIM) between the reenacted and source images and the average head pose and expression distance, APD and AED, between the target and the reenacted faces. On cross-subject reenactment, we expect the CSIM values to be lower that the one on self reenactment as the source and reenacted images are on different head pose. In Table~\ref{table:metrics_cross}, we report the quantitative comparisons extracted using $35$ random video pairs from the VoxCeleb1 test set. Our method is able to better transfer the target facial expression, while it remains competitive on head pose transfer. Although the CSIM metric is comparatively lower than HyperReenact~\cite{bounareli2023hyperreenact}, DPE~\cite{pang2023dpe} and Face2Face~\cite{yang2022face2face}, we assert that based on the qualitative results (Fig.~\ref{fig:comparisons}) our method is able to produce more realistic images, better capturing the source appearance and identity without producing visual artifacts.

        \begin{table}
       
            \begin{center}
            \begin{tabular}{|l|c|c|c|}
                \hline
                Method & CSIM & APD & AED \\
                \noalign{\hrule height 1.2pt}
                DaGAN~\cite{hong2022depth} & 0.56 & 2.1 & 18.2 \\
                Rome~\cite{khakhulin2022rome} & 0.63 & 1.2 & 13.3 \\
                Face2Face~\cite{yang2022face2face} & \underline{0.67} & 2.1 & 15.2 \\
                UniFace~\cite{xu2022designing} & 0.62 & 3.4 & 18.0 \\
                DPE~\cite{pang2023dpe} & \underline{0.67} & 3.7 & 16.0 \\
                \hline
                \hline
                StyleHeat~\cite{yin2022styleheat} & 0.57 & 3.5 & 15.5 \\
                FD~\cite{bounareli2022finding} & 0.49 & 1.7 & 14.6 \\
                HyperReenact~\cite{bounareli2023hyperreenact} & \textbf{0.68} & \textbf{0.5} & \underline{12.4} \\
                \hline
                \hline
                DiffusionRig~\cite{ding2023diffusionrig} & 0.34 & 1.3 & 14.8 \\
                Ours & 0.60 & \underline{1.0 } & \textbf{11.9} \\
                \hline
            \end{tabular}
            \end{center}
             \caption{Quantitative results on cross-subject reenactment using random video pairs from the test set of VoxCeleb1 dataset~\cite{Nagrani17}. For CSIM metric, higher is better, while for APD and AED, lower is better. We show the best and second best results in bold and underline.}\label{table:metrics_cross}
        \end{table}

         \begin{figure*}[h!]
            \centering
            \includegraphics[width=0.8\linewidth]{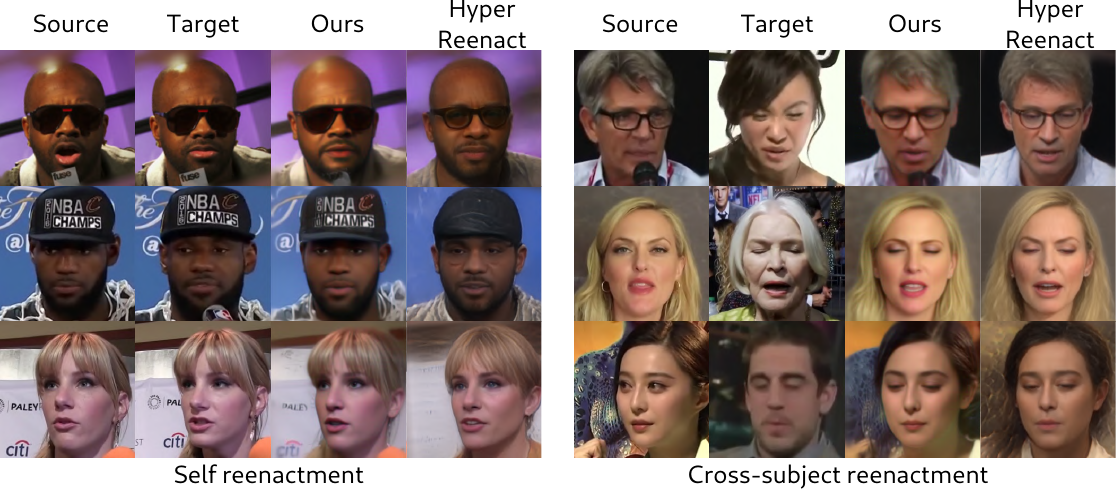}
            \caption{Comparisons against HyperReenact~\cite{bounareli2023hyperreenact} in both self~(left figure) and cross-subject~(right figure) reenactment. As clearly shown, our method achieves more accurate and realistic image reconstructions compared to HyperReenact.}
            \label{fig:comp_hyperreenact}
        \end{figure*}

        \noindent \textbf{Qualitative comparisons} We provide qualitative comparisons of the proposed method with state-of-the-art methods in Fig.~\ref{fig:comparisons}, on both self and cross-subject reenactment, along with additional comparisons and videos in the supplementary material. We also conducted a user study. Specifically, following the protocol of HyperReenact~\cite{bounareli2023hyperreenact}, we showed 20 random image pairs from VoxCeleb1 test set, 10 of self and 10 of cross-subject reenactment, to 35 users. We asked the users to select among the methods the one that synthesizes the best reenacted image in terms of identity preservation, facial pose transfer and image reconstruction quality. We report the results of the user study in Table~\ref{table:metrics_self} under the User Pref. column. As shown from both the results of the user study and from Fig.~\ref{fig:comparisons}, our approach is able to produce realistic and aesthetically pleasing images. Our method can accurately transfer the target facial pose, while also preserve the identity and appearance of the source faces both on self and on cross-subject reenactment tasks. Compared to the GAN-based methods, namely DaGAN~\cite{hong2022depth}, Rome~\cite{khakhulin2022rome}, Face2Face~\cite{yang2022face2face}, UniFace~\cite{xu2022designing} and DPE~\cite{pang2023dpe}, which achieve good quantitative results on CSIM metric, our method is able to more accurately transfer the target head pose and expression and also maintain the appearance of the source faces, without producing artifacts. We also perform on par on facial pose transfer with the StyleGAN2-based methods, namely StyleHeat~\cite{yin2022styleheat}, FD~\cite{bounareli2022finding} and HyperReenact~\cite{bounareli2023hyperreenact}. Regarding the image quality, StyleHeat~\cite{yin2022styleheat} and FD~\cite{bounareli2022finding} produce visual artifacts especially when there is difference on the head pose between the source and target faces, while our reenacted images are mostly artifact-free. Finally, while HyperReenact~\cite{bounareli2023hyperreenact} is able to produce visually pleasing images, it tends to over-smooth the faces and cannot accurately reconstruct details such as the hair styles, glasses or background elements, which are important for maintaining the overall perceptual quality of the generated images. A more detailed comparison with HyperReenact is illustrated in Fig.~\ref{fig:comp_hyperreenact}. As shown, while this method does not produce visual artifacts and successfully transfers the target facial pose, it fails in reconstructing important appearance details of the source images, e.g., glasses, hats, hair styles and background details. Finally, DiffusionRig~\cite{ding2023diffusionrig}, which is a diffusion-based method similar to ours, even though being able to transfer the target facial pose, it cannot produce realistic results using only one source frame, i.e. under one-shot settings. 

\subsection{Ablation studies}
\label{ssec:ablation}

    In this section, we present our ablation studies to assess how our training choices contribute to the final results. Specifically, we evaluate the contribution of (a) the pre-training process (``\textit{w/o} pre-training''), (b) the training protocol that involves both self reenactment and image reconstruction tasks (``\textit{w/o} batch split''), and (c) the joint optimization of the DDIM decoder with the proposed reenactment encoder $\mathcal{E}_r$ (``\textit{w/o} fine-tuning''). We provide both quantitative and qualitative comparisons in Table~\ref{table:ablation} and Fig.~\ref{fig:ablation}, respectively. In the supplementary material we also provide results of our final model with different steps $T$ to assess how the different steps contribute to the reconstruction quality of the generated images. 

    Regarding (a), we show results of our method without performing pre-training of the reenactment encoder, described in Sect.~\ref{ssec:training_protocol}. Instead the reenactment encoder is directly trained with the reconstruction losses. As shown in Table~\ref{table:ablation}, omitting the pre-training step (``\textit{w/o} pre-training'') results in less accurate transfer of head pose and expression. Additionally, our final model has better results on both PSNR and SSIM metrics, indicating that the reenacted images exhibit higher quality. Even though CSIM metric is relatively higher, as illustrated in Fig.~\ref{fig:ablation}, the reenacted images in 4th column (``\textit{w/o} pre-training'') have visual artifacts or fail to achieve the target facial pose. Consequently, this warm-up pre-training stage is crucial for enabling the reenactment encoder to better encode the target facial pose. 

    Regarding (b), we show that our training protocol that involves both self reenactment and image reconstruction task described in Sect.~\ref{ssec:training_protocol}, enables our model to better reconstruct the identity and appearance details of the source images. As shown in Table~\ref{table:ablation} and Fig.~\ref{fig:ablation}, ``\textit{w/o} batch split'' results in less accurate image reconstructions with visible deformations especially around the mouth area. Finally, for (c) we show that our model is able to produce realistic images even without the joint optimization of the DDIM decoder along with the reenactment encoder ``\textit{w/o} fine-tuning''. Nevertheless, by fine-tuning the DDIM decoder, our model is able to produce more accurate results in terms of the head pose and expression transfer as shown by the APD and AED metrics in Table~\ref{table:ablation}.

    \begin{figure}[h!]
        \centering
        \includegraphics[width=1.0\linewidth]{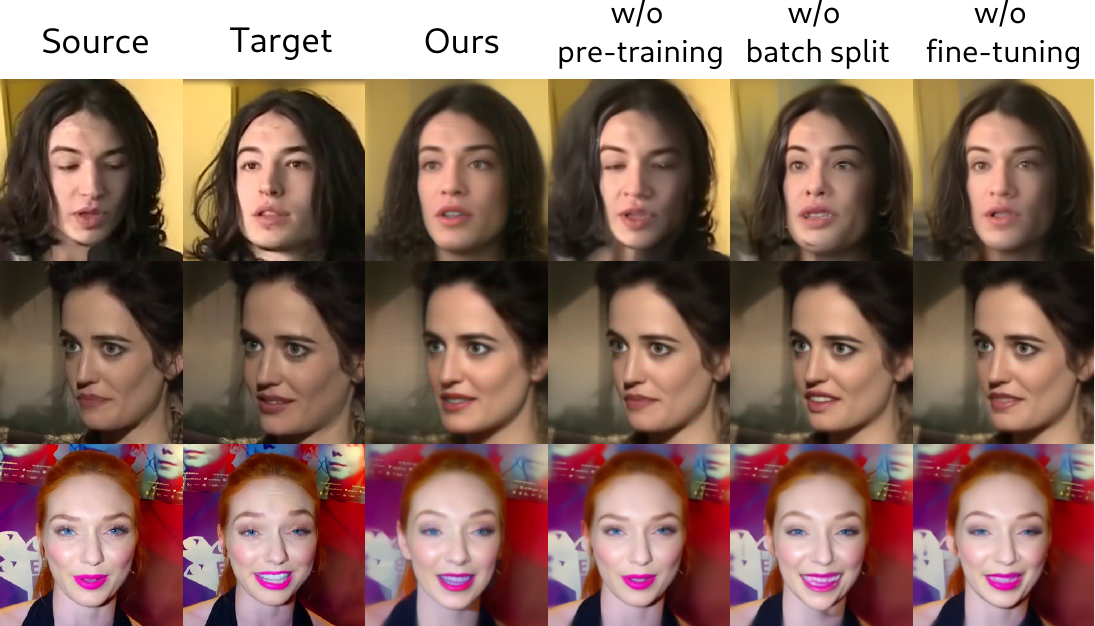}
        \caption{Qualitative comparisons of different training choices of our framework, i.e., without pre-training the reenactment encoder (``\textit{w/o} pre-training''), without the training protocol that involves both self reenactment and image reconstruction tasks (``\textit{w/o} batch split'') and without fine-tuning the DDIM decoder (``\textit{w/o} fine-tuning'').}
        \label{fig:ablation}
    \end{figure}
    
    \begin{table}[h]
    \small
   
    \begin{center}
    \begin{tabular}{|l|c|c|c|c|c|}
        \hline
        Method & CSIM & PSNR & SSIM & APD & AED \\
        \hline
        \textit{w/o} pre-training & \textbf{0.73} & 17.0 & 0.77 & 3.0 & 12.0   \\
        \textit{w/o} batch split & 0.62 & 18.7 & 0.80 & \underline{1.3} & 8.9   \\
        \textit{w/o} fine-tuning & 0.67 & \underline{19.0} & \underline{0.81} & \underline{1.3} & \underline{8.7} \\
        Ours &\underline{0.69} & \textbf{19.7 }& \textbf{0.83} & \textbf{ 0.8} & \textbf{7.2}\\
        \hline
    \end{tabular}
    \end{center}
     \caption{Quantitative comparisons of different training choices of our framework, i.e., without pre-training the reenactment encoder (``\textit{w/o} pre-training''), without the training protocol that involves both self reenactment and image reconstruction tasks (``\textit{w/o} batch split'') and without fine-tuning the DDIM decoder (``\textit{w/o} fine-tuning'')}\label{table:ablation}
    \end{table}


\section{Conclusions}
\label{sec:conclusions}

In this paper, we introduced a neural face reenactment framework, called \method, which is based on a pre-trained DPM. Our method employs a pre-trained diffusion autoencoder (DiffAE)~\cite{preechakul2022diffusionauto}, which consists of a semantic encoder and a DDIM sampler. The semantic encoder of DiffAE is able to encode an input image into a semantically meaningful code which can be used for image editing. Inspired by ContolNet~\cite{zhang2023controlnet}, we propose to control the semantic encoder of DiffAE using the target facial pose (reenactment encoder). Given a source image and a target facial pose, our reenactment encoder learns to predict the reenacted code that guides the generation process of DDIM to synthesize the reenacted image. We demonstrate that, in comparison to several state-of-the-art approaches, our method exhibits better or on-par reenactment performance, leading to facial images with less visual artifacts and deformations, showing that a pre-trained DDIM can effectively be adapted to the problem of neural face reenactment.

{\setlength{\parindent}{0cm}
\textbf{Acknowledgments:} This work was funded by UK Research and Innovation (UKRI) under the UK government’s Horizon Europe funding guarantee (grant No.~10099264) and by the European Union (under EC Horizon Europe grant agreement No.~101135800 (RAIDO)).
}

\appendix
\section{Supplementary Material}\label{supp}

In this supplementary material, we first provide in Sect.~\ref{sec:emoca}, details on how we extract i) the facial landmarks using the pre-trained EMOCA~\cite{danvevcek2022emoca} network and ii) the gaze position using an off-the-shelf pre-trained gaze estimation network~\cite{zhang2020eth}. In Sect.~\ref{sec:objectives}, we elaborate on the training objectives employed in the training phase of our pipeline (see Sect.~3.3 of the main paper). In Sect.~\ref{sec:steps}, we present an ablation study on the diffusion steps used in order to embed the real images into the stochastic subcode $\mathbf{x}_T$ and the steps used to reconstruct the reenacted images. In Sect.~\ref{sec:add_comparisons}, we show comparisons with two Stable Diffusion (SD) based methods~\cite{ye2023ip,wang2024instantid} and we provide additional qualitative comparisons on the VoxCeleb1 dataset~\cite{Nagrani17}, while also we present both quantitative and qualitative comparisons on the VoxCeleb2 dataset~\cite{Chung18b}. Furthermore, we demonstrate that our framework can generalize on other video datasets, namely HDTF~\cite{zhang2021flow} and VFHQ~\cite{xie2022vfhq}. In Sect.~\ref{sec:limitations}, we discuss the limitations of our method. Finally, along with this document, we provide a video file that includes video comparisons of our method with the state-of-the-art methods used in the main paper, as well as videos of our method on the HDTF~\cite{zhang2021flow} and VFHQ~\cite{xie2022vfhq} datasets.

\section{Facial landmarks with EMOCA~\cite{danvevcek2022emoca}}\label{sec:emoca}

    As discussed in the Sect.~3.2 of the main paper, the facial landmarks extracted using an off-the-shelf facial landmark detection method~\cite{bulat2017far} contain identity details, i.e., the facial shape of the input faces. As a result, when face reenactment methods use facial landmarks to drive the generation process, on the task of cross-subject face reenactment, identity leakage from the target faces into the reenacted ones is observed~\cite{zakharov2019few,zakharov2020fast}. To alleviate this issue, we use a 3D reconstruction method, namely EMOCA~\cite{danvevcek2022emoca}, which is based on 3D Morphable Models (3DMMs)~\cite{blanz1999morphable} and thus allows for reconstruction of the facial landmarks using the disentangled representations of the facial shape and pose. Specifically, a 3D facial shape $\textbf{s}\in\mathbb{R}^{3N}$ can be parameterized as:
    \begin{equation}\label{eq:3dmm}
        \textbf{s} = \bar{\textbf{s}} + \textbf{S}_{i}\mathbf{p}_{i} +  \textbf{S}_{\theta}\mathbf{p}_{\theta} + \textbf{S}_{e}\mathbf{p}_{e},
    \end{equation}
    where $\bar{\textbf{s}}$ denotes the mean 3D facial shape, $\textbf{S}_{i}\in\mathbb{R}^{3N\times m_{i}}$, $\textbf{S}_{\theta}\in\mathbb{R}^{3N\times m_{\theta}}$, and $\textbf{S}_{e}\in\mathbb{R}^{3N\times m_{e}}$ denote respectively the PCA bases for identity, head pose orientation, and expression, and $\mathbf{p}_{i}$, $\mathbf{p}_{\theta}$, and $\mathbf{p}_{e}$ denote respectively the corresponding identity, head pose orientation, and expression coefficients.
    
\begin{figure}[t]
        \centering
        \includegraphics[width=1.0\linewidth]{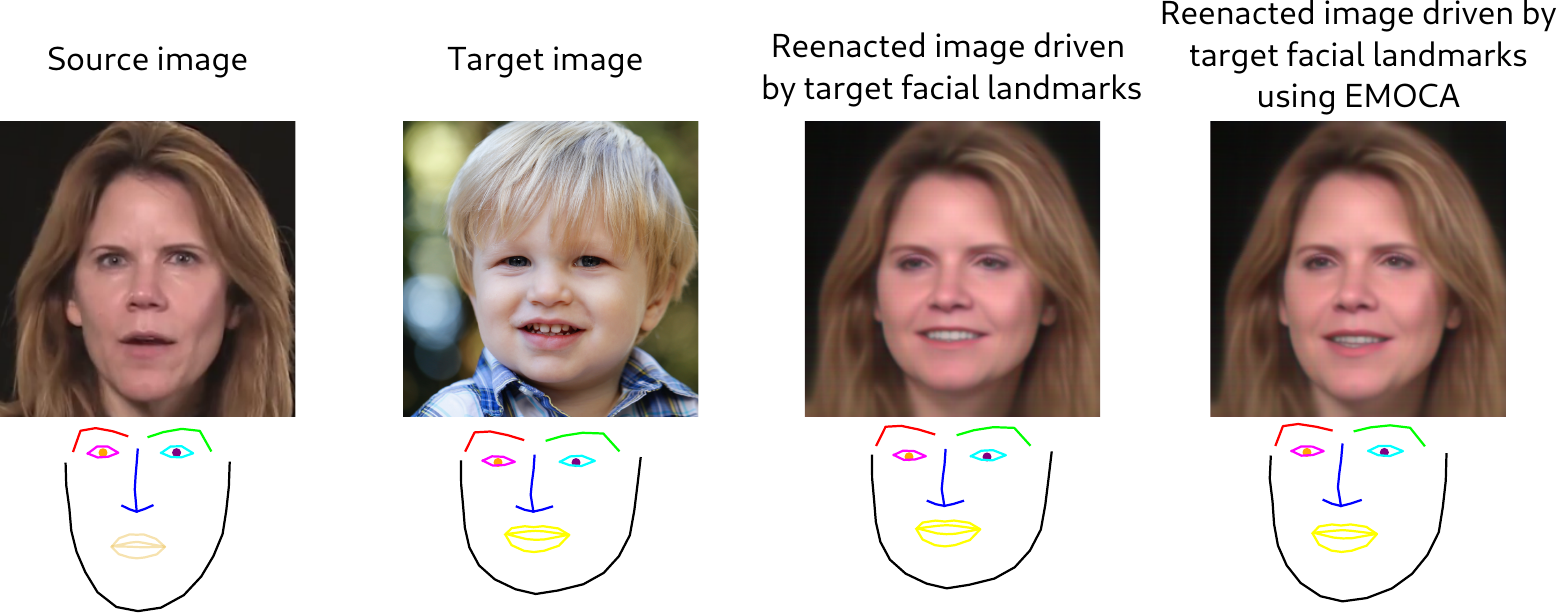}
        \caption{Example of cross-subject face reenactment driven by the target facial landmarks extracted from the off-the-shelf facial landmark estimation network of~\cite{bulat2017far} and by the reconstructed facial landmarks extracted from EMOCA~\cite{danvevcek2022emoca} (as described in Eq.~\ref{suppeq:3dmm}). Clearly, when employing the target facial landmarks from~\cite{bulat2017far}, the reenacted image has the facial shape of the target image (identity leakage), which is avoided using the adopted EMOCA~\cite{danvevcek2022emoca} network.}
        \label{fig:landmarks_emoca}
    \end{figure}

    Having this representation of the 3D facial shape and considering as input a pair of source and target images from different identities, we can calculate the ground-truth target facial shape $\textbf{s}^t$. Specifically, $\textbf{s}^t$ is obtained using the identity coefficients $\mathbf{p}_{i}^s$ of the source image and the head pose and expression coefficients $\mathbf{p}_{\theta}^t, \mathbf{p}_{e}^t$ of the target image as follows:
    \begin{equation}\label{suppeq:3dmm}
        \textbf{s}^t = \bar{\textbf{s}} + \textbf{S}_{i}\mathbf{p}_{i}^s +  \textbf{S}_{\theta}\mathbf{p}_{\theta}^t + \textbf{S}_{e}\mathbf{p}_{e}^t.
    \end{equation}
    Hence, the corresponding target facial landmarks contain the head pose and expression of the target image and the facial shape of the source image, overcoming the identity leakage from the target image to the reenacted ones. An example of cross-subject reenactment is illustrated in Fig.~\ref{fig:landmarks_emoca}. The first and second columns show respectively the source and the target images which have large facial shape difference; i.e., the source image has long facial shape, while the target image has round shape. In the third column, we present the reenacted image generated using the target facial landmarks extracted from a pre-trained landmark estimation network~\cite{bulat2017far}. The last column shows the reenacted images generated using the reconstructed facial landmarks extracted from EMOCA~\cite{danvevcek2022emoca} as described in (\ref{suppeq:3dmm}) above. Clearly, when employing the target facial landmarks from~\cite{bulat2017far}, the reenacted image inherits the facial shape of the target image, which results in missing important identity details from the source image.

    Finally, in order to control the gaze direction on the reenacted faces, as discussed in Sect.~3.2 of the main paper, along with the standard facial landmarks, we further used an additional landmark point that corresponds to the gaze position. Specifically, we estimated the gaze direction angles $\alpha, \beta$ using the pre-trained gaze estimation network of~\cite{zhang2020eth}, and by using the landmark points that correspond to the eyes, we calculated the width $w_e$, height $h_e$, and the center $c_e$ of each eye, and the gaze position $(g_x, g_y)$ as follows:
    \begin{equation}
        \begin{aligned}
          & g_x = c_e -\frac{w_e}{2} \cdot \sin(\beta) \cdot \cos(\alpha), \\
          & g_y = c_e -\frac{h_e}{2} \cdot \sin(\alpha).
        \end{aligned}
    \end{equation}

\section{Training objectives}\label{sec:objectives}

\begin{figure}[t]
        \centering
        \includegraphics[width=1.0\linewidth]{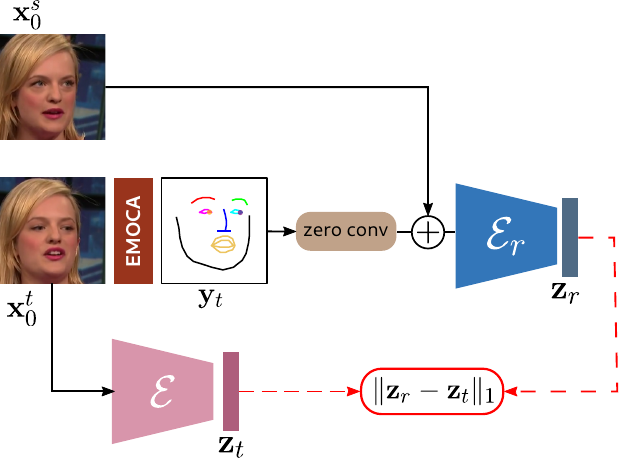}
        \caption{\textbf{Pre-training process:} We pre-train the reenactment encoder $\mathcal{E}_r$ under the self reenactment setting using only the $\ell_1$ loss between the reenacted code $\mathbf{z}_r$ and the target code $\mathbf{z}_t$ given by the semantic encoder $\mathcal{E}$ of DiffAE~\cite{preechakul2022diffusionauto}.}
        \label{supfig:train_1}
    \end{figure}

    As discussed in the Sect.~3.3 of the main paper, we propose to train our method in two stages; a pre-training stage and a main training stage. Specifically, as illustrated in Fig.~\ref{supfig:train_1}, we pre-train the reenactment encoder $\mathcal{E}_r$ under the self reenactment setting, where we only use the $\ell_1$ loss, $ \mathcal{L}_z =\lVert \mathbf{z}_r - \mathbf{z}_t \rVert_1$, between the reenacted code $\mathbf{z}_r$ predicted by the trainable model $\mathcal{E}_r$ and the target code $\mathbf{z}_t$ predicted by the semantic encoder $\mathcal{E}$ of DiffAE~\cite{preechakul2022diffusionauto}. We then continue training the reenactment encoder $\mathcal{E}_r$ (main training stage) by incorporating the DDIM sampler~\cite{Song2021ddim} as a generative module to synthesize the reenacted images (as illustrated in Fig.~2 of the main paper). Our training objective in this stage combines both reconstruction and poss losses to ensure that the reenacted image will depict the source identity in the target facial pose (head orientation and facial expressions). In this section we present in detail the adopted objectives used during training.

    \noindent \textbf{Pixel-wise loss $\mathcal{L}_{pix}$} The pixel-wise loss $\mathcal{L}_{pix}$ is defined as the $\ell_2$-distance between the target $\mathbf{x}_0^t$ and the reenacted images $\mathbf{x}_0^r$:
    \begin{equation}
        \mathcal{L}_{pix} = \lVert \mathbf{x}_0^t - \mathbf{x}_0^r \rVert_2
    \end{equation}

    \noindent \textbf{Perceptual loss $\mathcal{L}_{per}$} Instead of relying only on the pixel-wise loss, we calculate the perceptual loss $\mathcal{L}_{per}$, which is defined as the sum of $\ell_2$-distances between the intermediate features of a pre-trained VGG19 network $\phi$~\cite{johnson2016perceptual}:
    \begin{equation}
        \mathcal{L}_{per} = \sum_{l=1}^{N} \lVert \phi_l(\mathbf{x}_0^t) - \phi_l(\mathbf{x}_0^r) \rVert_2,
    \end{equation}
    where $N$ is the number of layers used from the VGG19 network to compute the perceptual loss.

    \noindent \textbf{Identity loss $\mathcal{L}_{id}$} In order to preserve the identity characteristics of the input faces, we calculate the identity loss $\mathcal{L}_{id}$, which is defined as the cosine similarity of the features extracted from the pre-trained ArcFace~\cite{deng2019arcface} network $\mathcal{F}$:
    \begin{equation}
        \mathcal{L}_{id} = 1 - \frac{\mathcal{F}(\mathbf{x}_0^t)\cdot\mathcal{F}(\mathbf{x}_0^r)}{ \|\mathcal{F}(\mathbf{x}_0^t)\| \|\mathcal{F}(\mathbf{x}_0^r)\| }
    \end{equation}

    \noindent \textbf{Background loss $\mathcal{L}_{bg}$} In order to enhance the background preservation, we calculate the background loss $\mathcal{L}_{bg}$ by using the off-the-shelf pre-trained face segmentation network of~\cite{yu2021bisenet} to create a mask $m_f$ around the face area so as to keep only the background and hair areas. We then calculate the background loss as:
    \begin{equation}
        \mathcal{L}_{bg} = \lVert \mathbf{x}_0^t \odot m_f^t  -  \mathbf{x}_0^r \odot m_f^r \rVert_2,
    \end{equation}
    where $m_f^t$, $m_f^r$ are the corresponding masks of the target and the reenacted images.

    \noindent \textbf{Style loss $\mathcal{L}_{st}$} To improve the preservation of the input facial appearance, we incorporate the style loss $\mathcal{L}_{st}$, similarly to~\cite{barattin2023attribute}. Specifically, we use features from the pre-trained FaRL~\cite{zheng2022general} ViT-based image encoder, which is trained on image and text pairs to learn meaningful feature representations of facial images. We calculate the style loss as the $\ell_2$-distance of the $512$-dimensional feature vectors extracted from FaRL encoder $\mathcal{E}_{farl}$ as follows:
    \begin{equation}
        \mathcal{L}_{st} = \lVert \mathcal{E}_{farl}(\mathbf{x}_0^t) - \mathcal{E}_{farl}(\mathbf{x}_0^r ) \rVert_2.
    \end{equation}

    \noindent \textbf{Gaze loss $\mathcal{L}_{g}$} To control the gaze direction on the reenacted images, we calculate the gaze loss, which is defined as the $\ell_1$-distance of the gaze direction, i.e., the gaze angles, between the target and the reenacted images extracted from the pre-trained gaze estimation network of~\cite{zhang2020eth}, as follows:
    \begin{equation}
        \mathcal{L}_{g} = \lVert g(\mathbf{x}_0^t)) - g(\mathbf{x}_0^r) \rVert_1.
    \end{equation}

    \noindent \textbf{Shape loss $\mathcal{L}_{shape}$} In order to transfer the target facial pose defined as the head pose orientation and the facial expressions we calculate the shape loss, which is defined as the $\ell_1$-distance between the target and the reenacted 2D facial landmarks extracted from EMOCA~\cite{danvevcek2022emoca}:
    \begin{equation}
        \mathcal{L}_{shape} = \lVert l^t - l^r \rVert_1.
    \end{equation}

        \noindent \textbf{Head pose loss $\mathcal{L}_{hp}$} To improve the head pose transfer, we calculate the mean $\ell_1$-distance of the head pose orientation, i.e., the three Euler angles $a$, yaw, pitch and roll, between the target and the reenacted images:
    \begin{equation}
        \mathcal{L}_{hp} = \lVert a^t - a^r \rVert_1,
    \end{equation}

\section{Ablation study on diffusion steps $T$}\label{sec:steps}
In this section we report results of our method using different steps $T$ required to decode the reenacted images and steps $T_{\mathbf{x}_T}$ required to encode the source images into the subcode $\mathbf{x}_T$ (please see Sect.~3.1 of the main paper and~\cite{preechakul2022diffusionauto}). Specifically, we asses the quality of the reenacted images while varying the different steps used in the diffusion process. Hence, we report only the reconstruction quality metrics, namely PSNR, SSIM, CSIM, and LPIPS, as there are no changes on the facial pose metrics (NME, ARD, and AED). We also report the inference time required to reenact 20 frames. We note that for all the results reported in the main paper and in the supplementary material we use $T_{\mathbf{x}_T} = 50$ and $T=20$.  

\begin{figure}[t]
    \centering
    \includegraphics[width=1.0\linewidth]{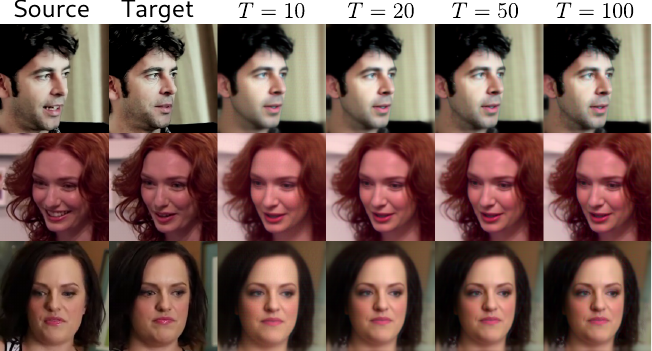}
    \caption{Qualitative results of our model using different steps $T$ to generate the reenacted images. We note that in the experimental evaluation of our method (Sect.~4.2 of the main paper), we use $T_{\mathbf{x}_T} = 50$ and $T=20$.}
    \label{supfig:ablation_steps_T}
\end{figure}

    In Table~\ref{supptable:ablation_stepsT} we show comparisons of our model using $T_{\mathbf{x}_T} = 50$ while varying the steps $T$ required to generate the reenacted images. We note that our model achieves the best results when $T=20$ and $T=50$ with minor changes when $T=100$. Additionally, as illustrated in Fig.~\ref{supfig:ablation_steps_T}, the reenacted images using $T=10$ are more noisy compared to the ones with larger steps. We argue that taking into consideration both the reconstruction metrics and the inference time, the best results are obtained when $T=20$.

\begin{table}
   
    \begin{center}
    \small
    \begin{tabular}{|c|c|c|c|c|c|}
        \hline    
        \multirow{2}{*}{$T_{\mathbf{x}_T} = 50$} & \multirow{2}{*}{PSNR} &  \multirow{2}{*}{SSIM} &  \multirow{2}{*}{CSIM} &  \multirow{2}{*}{LPIPS} & \multirow{2}{*}{\parbox{1cm}{\centering Inf. \\ time}}\\
        &  &  &  &  &  \\
        \hline
        $T=10$ & 19.6 & 0.82 & 0.66 & 0.27 & 24.9 \\
        $T=20$ & 19.7 & 0.83 & 0.69 & 0.24 & 34.0\\
        $T=50$ & 19.7 & 0.83 & 0.69 & 0.24 & 82.8\\
        $T=100$ & 19.6 & 0.82 & 0.68 & 0.25 & 162.0 \\
        \hline
    \end{tabular}
     \caption{Quantitative results of our model using different steps $T$ to generate the reenacted images. We note that in the experimental evaluation of our method (Sect.~4.2 of the main paper), we use $T_{\mathbf{x}_T} = 50$ and $T=20$.}\label{supptable:ablation_stepsT}
    \end{center}
    
\end{table}

\begin{table}
   
    \begin{center}
    \small
    \begin{tabular}{|c|c|c|c|c|c|}
        \hline
       \multirow{2}{*}{$T = 20$} & \multirow{2}{*}{PSNR} &  \multirow{2}{*}{SSIM} &  \multirow{2}{*}{CSIM} &  \multirow{2}{*}{LPIPS} & \multirow{2}{*}{\parbox{1cm}{\centering Inf. \\ time}}\\
       
        &  &  &  &  &  \\
        \hline
        random $\mathbf{x}_T$ & 19.0  & 0.80 & 0.47 & 0.33  & 30.2 \\
        $T_{\mathbf{x}_T}=20$ & 19.6  & 0.82 &  0.68 & 0.25 & 32.4\\
        $T_{\mathbf{x}_T}=50$ & 19.7 & 0.83 & 0.69 & 0.24 & 34.0 \\
        $T_{\mathbf{x}_T}=100$ & 19.7  & 0.83  & 0.69  & 0.24 & 42.0 \\
        $T_{\mathbf{x}_T}=250$ & 19.7  & 0.83 & 0.69 & 0.24 & 54.0\\
        \hline
    \end{tabular}
    \end{center}
     \caption{Quantitative results of our model using different steps $T_{\mathbf{x}_T}$ to encode the source image into the subcode $\mathbf{x}_T$. We also report results when the subcode $\mathbf{x}_T$ is random, sampled from $\mathcal{N}(0, \mathbf{I})$. We note that we use $T=20$ steps to generate the reenacted images.}\label{supptable:ablation_stepsT_xt}
\end{table}

\begin{figure}[t]
    \centering
    \includegraphics[width=1.0\linewidth]{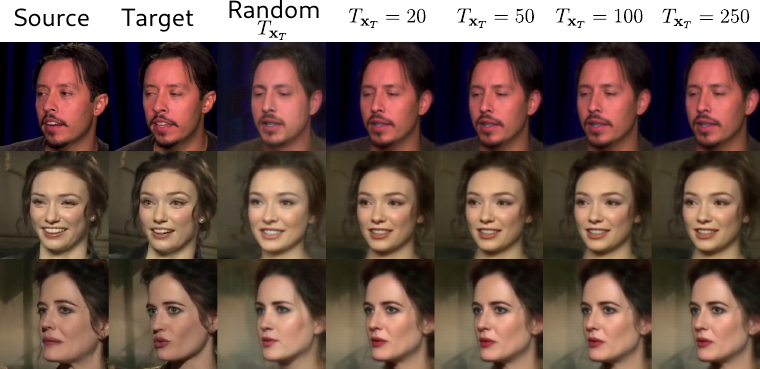}
    \caption{Qualitative results of our model using different steps $T$ to generate the reenacted images. We note that in the experimental evaluation of our method (Sect.~4.2 of the main paper), we use $T_{\mathbf{x}_T} = 50$ and $T=20$.}
    \label{supfig:ablation_steps_xt}
\end{figure}

    Finally, in Table~\ref{supptable:ablation_stepsT_xt} we report the results of our method using $T = 20$ while varying the steps $T_{\mathbf{x}_T}$ required to encoder the stochastic subcode $\mathbf{x}_T$. We also show results when $\mathbf{x}_T$ is random, sampled from $\mathcal{N}(0, \mathbf{I})$. We note that we calculate the stochastic subcode of the source image for each video. Fig.~\ref{supfig:ablation_steps_xt} illustrates the qualitative comparisons while varying the steps $T_{\mathbf{x}_T}$. As shown, when using a random subcode $\mathbf{x}_T$ the reenacted images are not as good as when we use the source images to encode the stochastic subcode. Taking into consideration the inference time, the quantitative and qualitative comparisons, we achieve the best results when $T_{\mathbf{x}_T} = 50$.

\section{Experiments}\label{sec:add_comparisons}

\subsection{Analysis on identity preservation}

\begin{figure}[h!]
    \centering
    \includegraphics[width=1.0\linewidth]{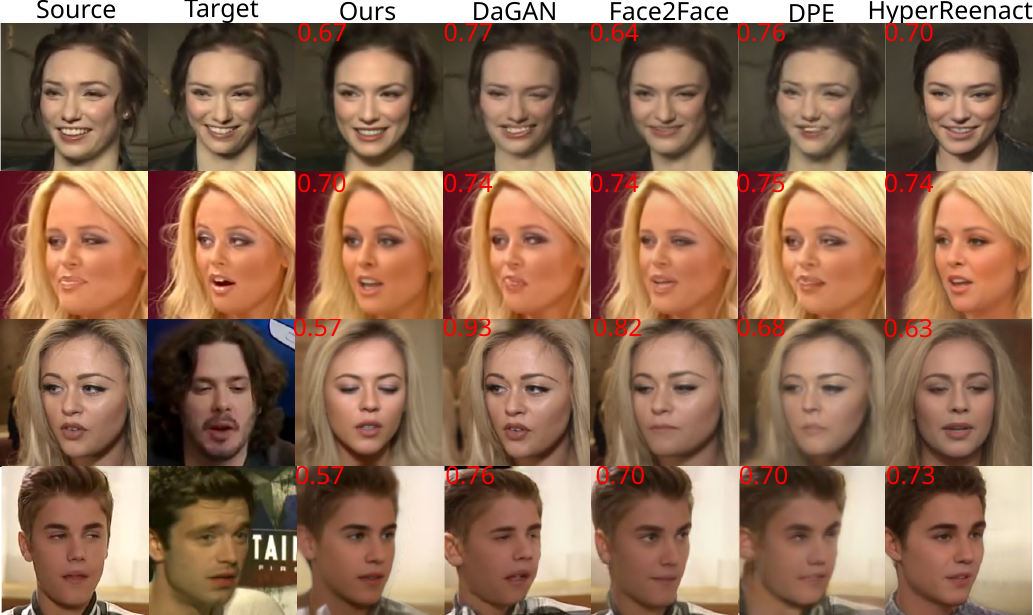}
    \caption{Visual comparisons on self (first two rows) and cross-subject (last two rows) reenactment against methods that achieve high CSIM score, namely DaGAN~\cite{hong2022depth}, Face2Face~\cite{yang2022face2face}, DPE~\cite{pang2023dpe} and HyperReenact~\cite{bounareli2023hyperreenact}. We also report for each method the corresponding CSIM score for the depicted samples. As shown a higher CSIM score does not necessarily imply better reenactment results.}
    \label{supfig:csim}
\end{figure}

Neural face reenactment is a complex task that requires several quantitative and qualitative results in order to fairly evaluate the various methods. Thus, we provide several metrics in order to evaluate the reconstruction quality and identity preservation of the reenacted images, i.e., PNSR, SSIM, L1, LPIPS and CSIM and additional metrics to assess the head pose and expression transfer, i.e., NME, APD and AED. Along with the quantitative metrics we also conduct a user study to assess the human quality perception. Our approach surpasses all other methods on PSNR, SSIM and L1 metrics, demonstrating our higher reconstruction ability. Nevertheless, on identity preservation, i.e., CSIM metric, although our method remains competitive, DaGAN~\cite{hong2022depth}, Face2Face~\cite{yang2022face2face}, DPE~\cite{pang2023dpe} and HyperReenact~\cite{bounareli2023hyperreenact} have higher CSIM scores. We argue that the CSIM metric is not the most suitable to evaluate the reenactment methods and as a result a higher CSIM score does not necessarily imply better reenacted images. In order to support our argument we show in Fig.~\ref{supfig:csim} visual comparisons between the methods that achieve higher CSIM scores. As shown the generated images of DaGAN~\cite{hong2022depth}, Face2Face~\cite{yang2022face2face} and DPE~\cite{pang2023dpe} have visual artifacts and deformations, however the corresponding CSIM scores are high, indicating that CSIM is invariant to such distortions. Additionally, HyperReenact~\cite{bounareli2023hyperreenact} over-smooths the generated images and misses important appearance details.

\subsection{Additional comparisons}

\begin{figure}[h!]
    \centering
    \includegraphics[width=1.0\linewidth]{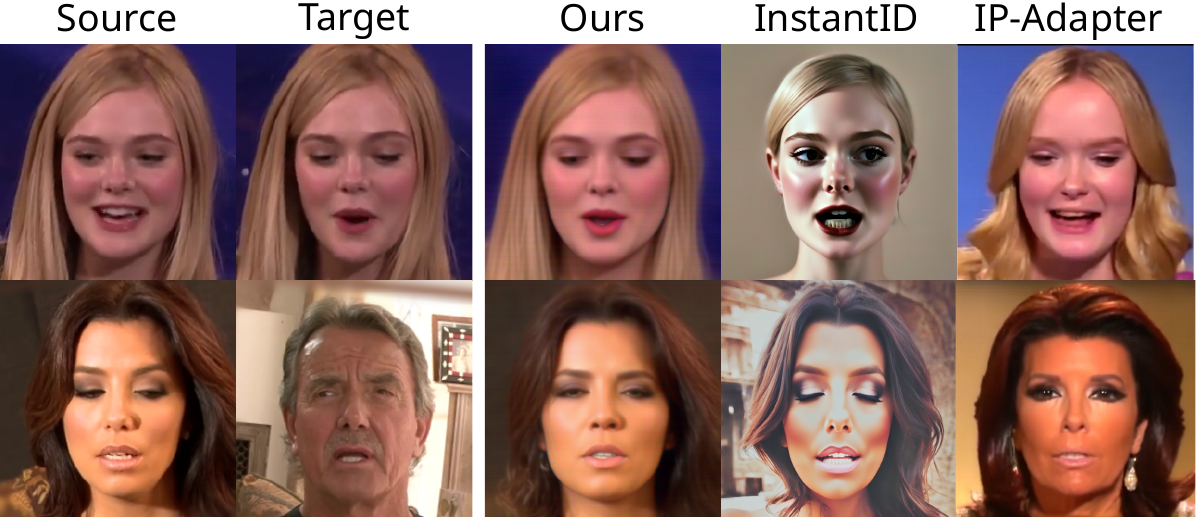}
    \caption{Comparisons on self and cross-subject reenactment against the SD-based methods InstantID~\cite{wang2024instantid} and IP-Adapter~\cite{ye2023ip}. Clearly, these two methods are not able to preserve both the identity and appearance of the source images.}
    \label{fig:sd_comparisons}
\end{figure}

We provide comparisons with two recent methods, namely InstantID~\cite{wang2024instantid} and IP-Adapter~\cite{ye2023ip} that condition SD models~\cite{rombach2022high} to generate controllable images that maintain the style of the input faces. As shown in Fig.~\ref{fig:sd_comparisons}, such methods are not able to preserve the identity and appearance of the source faces and accurately transfer the target head pose and expression, making them impractical for the task of face reenactment. In contrast, our method leverages the high quality generation of Diffusion Probabilistic Models (DPMs) and produces accurate reconstructions.

We provide additional results and comparisons on the VoxCeleb1~\cite{Nagrani17}, VoxCeleb2~\cite{Chung18b}, HDTF~\cite{zhang2021flow} and VFHQ~\cite{xie2022vfhq} datasets. Specifically, in Figs.~\ref{supfig:comparisons_self_vox1},~\ref{supfig:comparisons_cross_vox1} we provide qualitative comparisons on self and cross-subject reenactment, respectively, using the test set of the VoxCeleb1 dataset. Additionally, in Table~\ref{supptable:benchmark} we report the quantitative results on the small benchmark dataset released by the authors of HyperReenact~\cite{bounareli2023hyperreenact}. This benchmark dataset contains image pairs drawn from the VoxCeleb1 test set where the head pose distance between the source and target frames is larger than $15^{\circ}$. We note that our method remains competitive on the challenging task of extreme head pose variations, both on facial pose transfer and on image reconstruction quality. As illustrated in Fig.~\ref{supfig:comparisons_benchmark}, the majority of the state-of-the-art methods produce many deformations on the generated images. In contrast, our method can effectively preserve the source appearance/identity and transfer the target facial pose (head orientation and facial expression) without producing visual artifacts and deformations.

\begin{table*}
   
    \begin{center}
    \begin{tabular}{|l|c|c|c|c|c|c|c|c|}
        \hline
        Method & PSNR & SSIM & CSIM & LPIPS & L1 &  NME &  APD & AED \\
        \noalign{\hrule height 1.2pt}
        DAGAN~\cite{hong2022depth} & 16.3 & 0.76 & 0.44 & \underline{0.32} & 0.21 &  22.6 & 3.4 & 12.7 \\
        Rome~\cite{khakhulin2022rome} & 6.9 & 0.71 & \underline{0.53} & 0.48 & 0.67 &  \underline{14.6} & 1.2 & 9.2 \\
        Face2Face~\cite{yang2022face2face} & 14.0 & 0.72 & 0.39 & 0.49 & 0.30 &  21.9 & 3.8 & 12.4 \\
        UniFace~\cite{xu2022designing} & 14.5 & 0.68 & 0.27 & 0.37 & 0.27 &  26.1 & 3.8 & 14.6  \\
        DPE~\cite{pang2023dpe} & 16.3 & 0.75 & 0.45 & 0.37 & 0.22 &  26.7 & 4.7 & 12.9  \\
        \hline
        \hline
        StyleHeat~\cite{yin2022styleheat} &  14.5 & 0.77 & 0.45 & 0.40 & 0.28 & 32.4&  8.6 & 13.9  \\
        FD~\cite{bounareli2022finding} & 15.8 & 0.74 & 0.37 & 0.35 & 0.23 &  23.2 & 2.7 & 12.4  \\
        HyperReenact~\cite{bounareli2023hyperreenact} & \underline{17.0} & \underline{0.79} & \textbf{0.60} & \textbf{0.28} & \textbf{0.18} &  \textbf{13.8} & \textbf{0.6} & \underline{8.6} \\
        \hline
        \hline
        DiffusionRig~\cite{ding2023diffusionrig} & 13.3 & 0.71 & 0.24 & 0.52 & 0.34 & 54.0 & 1.7 & 12.5 \\
        Ours & \textbf{17.5} & \textbf{0.83} & 0.49 & 0.34 & \underline{0.19} &  \underline{14.6} & \underline{1.0} & \textbf{8.0}   \\
        \hline
    \end{tabular}
    \end{center}
     \caption{Quantitative results on the small benchmark dataset with large head pose variations released by the authors of HyperReenact~\cite{bounareli2023hyperreenact}. For PSNR, SSIM, CSIM metrics, higher is better, while for the rest of the metrics lower is better. We note that the best and second best results are shown in bold and underline respectively.}\label{supptable:benchmark}
\end{table*}

    In order to evaluate the temporal coherence of the reenacted videos, following~\cite{tzaban2022stitch}, we calculate the temporally-local (TL-ID) and temporally-global (TG-ID) identity preservation metrics. Specifically, those metrics assess the local and the global consistency of the identity similarity between consecutive frames and across all frames of a reenacted video compared to the real one. For both metrics, a higher score indicates higher temporal consistency. As shown in Table~\ref{supptable:temporal}, our method achieves high temporal consistency, especially compared to DiffusionRig~\cite{ding2023diffusionrig} that is also a diffusion-based method. Finally, in Table~\ref{supptable:temporal} we report the inference time of each method in order to reenact a video with 200 frames. We note that the diffusion-based methods, namely ours and DiffusionRig, are not as fast as the GAN-based method, which is an expected limitation of diffusion models. Nevertheless, our method is faster compared to DiffusionRig.   

\begin{table}[t]
    
    \begin{center}
    \begin{tabular}{|l|c|c|c|}
        \hline
        Method & TL-ID & TG-ID & Inf. time (sec)\\
        \hline
        DAGAN~\cite{hong2022depth} & 0.992 & 0.952 & \textbf{11}\\
        Rome~\cite{khakhulin2022rome} & 0.993 & \textbf{1.000} & 100 \\
        Face2Face~\cite{yang2022face2face} & 0.985 & 0.981 & 140 \\
        UniFace~\cite{xu2022designing} & 0.986 & 0.788 & 31\\
        DPE~\cite{pang2023dpe} & \textbf{1.000}  & \underline{0.995} & 30\\
        \hline
        \hline
        StyleHeat~\cite{yin2022styleheat} & \textbf{1.000}  & 0.936 & \underline{25}\\
        FD~\cite{bounareli2022finding} & \underline{0.998} & 0.834 & 40 \\
        HyperReenact~\cite{bounareli2023hyperreenact} & \textbf{1.000} & 1.0 & 40  \\
        \hline
        \hline
        DiffusionRig~\cite{ding2023diffusionrig} & 0.87  & 0.52 & 450 \\
        Ours & \textbf{1.000}  & 0.950 & 340 \\
        \hline
    \end{tabular}
    \end{center}
    \caption{Quantitative comparisons of the temporal coherence (TL-ID, TG-ID) and inference time.}\label{supptable:temporal}
\end{table}

Additionally, we provide quantitative and qualitative comparisons on the VoxCeleb2~\cite{Chung18b} test set. Specifically, in Table~\ref{supptable:vox2_self} we report the quantitative results on self reenactment. Additionally, in Figs.~\ref{supfig:comparisons_self_vox2},~\ref{supfig:comparisons_cross_vox2} we present qualitative comparisons on self and cross-subject reenactment, respectively. As shown in Table~\ref{supptable:vox2_self}, our method outperforms all methods on PNSR and SSIM metrics, while we also achieve best results on head pose and expression transfer metrics (NME, APD and AED). As shown from the qualitative comparisons in Figs.~\ref{supfig:comparisons_self_vox2},~\ref{supfig:comparisons_cross_vox2}, our method can produce accurate reconstructions. Compared to DaGAN~\cite{hong2022depth}, Face2Face~\cite{yang2022face2face}, and DPE~\cite{pang2023dpe}, which achieve good performance on the CSIM metric, our reenacted images are artifact-free and more realistic (e.g., see rows 3 and 6 of Fig.~\ref{supfig:comparisons_self_vox2}). Notably, those methods when evaluated in the task of cross-subject reenactment, Fig.~\ref{supfig:comparisons_cross_vox2}, which is more challenging compared to self reenactment, produce many deformations on the generated images, rendering their results unrealistic. Finally, HyperReenact~\cite{bounareli2023hyperreenact}, which has competitive quantitative metrics, as shown in Fig.~\ref{supfig:comparisons_self_vox2} (5th,6th, and 7th rows) and Fig.~\ref{supfig:comparisons_cross_vox2} (7th row), cannot faithfully reconstruct certain appearance details (e.g., glasses and hats) of the source images. In contrast, our method can effectively reconstruct those accessories, leading to more accurate and natural reenacted images. Finally, in Figs.~\ref{suppfig:comparisons_hdtf},~\ref{suppfig:comparisons_vfhq}, we present qualitative results of our method on HDTF~\cite{zhang2021flow} and VFHQ~\cite{xie2022vfhq} datasets. Clearly, our method can generalize well on other video datasets and generate realistic images that resemble the real ones both in self and cross-subject reenactment.

\begin{table*}
    
    \begin{center}
    \begin{tabular}{|l|c|c|c|c|c|c|c|c|}
        \hline
        Method &  PSNR & SSIM & CSIM & LPIPS & L1 & NME &  APD  & AED \\
        \noalign{\hrule height 1.2pt}
        DAGAN~\cite{hong2022depth} & 16.8 & 0.74 &\underline{0.71} & 0.23 & 0.18 &  27.3 & 4.5 & 13.0\\
        Rome~\cite{khakhulin2022rome} & 7.6 & 0.73 & 0.64 & 0.46 & 0.63 &  19.2 & 1.8 & 9.7\\
        Face2Face~\cite{yang2022face2face} & 17.6 & 0.77 & \underline{0.71} & 0.23 & \underline{0.17} &  16.7 & 1.5 & 10.5 \\
        UniFace~\cite{xu2022designing} & \underline{18.1} & 0.79 & 0.70 & \textbf{0.20} & \textbf{0.16}  & 18.1 & 2.0 & 12.0  \\
        DPE~\cite{pang2023dpe} & 18.0 & 0.77 & \textbf{0.72} & \underline{0.21} & 0.19 &  18.5 & 3.5 & 12.6    \\
        \hline
        \hline
        StyleHeat~\cite{yin2022styleheat} & 17.2 & 0.80 & 0.56 & 0.29 & 0.19 &  20.3 & 3.5 &  12.6 \\
        FD~\cite{bounareli2022finding} & 17.0 & 0.77 & 0.64 & 0.25 & 0.18 &  \underline{16.2} & 1.3 &  9.8 \\
        HyperReenact~\cite{bounareli2023hyperreenact} & 17.3 & 0.80 & 0.67 & 0.29 & 0.19 &  \textbf{13.6} & \textbf{0.7} & \underline{8.7}\\
        \hline
        \hline
        DiffusionRig~\cite{ding2023diffusionrig} & 14.7 & 0.75 & 0.30 & 0.41  & 0.27 & 24.8  & 1.2  & 11.6 \\
        Ours & \textbf{18.4} & \textbf{0.84} & 0.64 & 0.26 & \textbf{0.16}  & \textbf{ 13.6} & \underline{0.8} & \textbf{7.5} \\
        \hline
    \end{tabular}
    \end{center}
    \caption{Quantitative results on self reenactment on the VoxCeleb2 dataset~\cite{Chung18b}. For PSNR, SSIM, CSIM metrics, higher is better, while for the rest of the metrics lower is better. We note that the best and second best results are shown in bold and underline respectively.}\label{supptable:vox2_self}
\end{table*}

\begin{figure}[h!]
        \centering
        \includegraphics[width=0.8\linewidth]{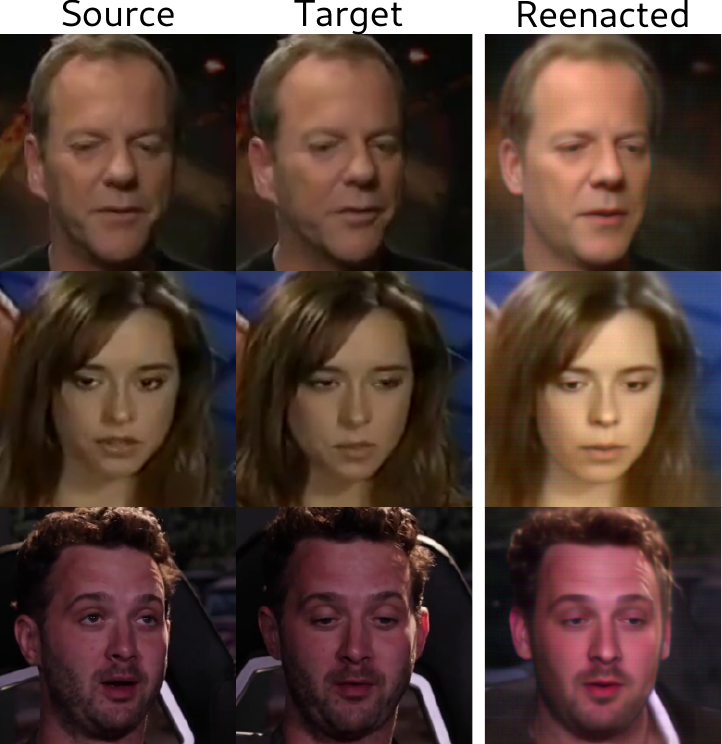}
        \caption{We observe that in certain cases, especially when the input videos are darker and low-resolution, the reenacted images are brighter than the real ones.}
        \label{supfig:limitations}
\end{figure}

\begin{figure*}[t]
    \centering
    \includegraphics[width=1.0\textwidth]{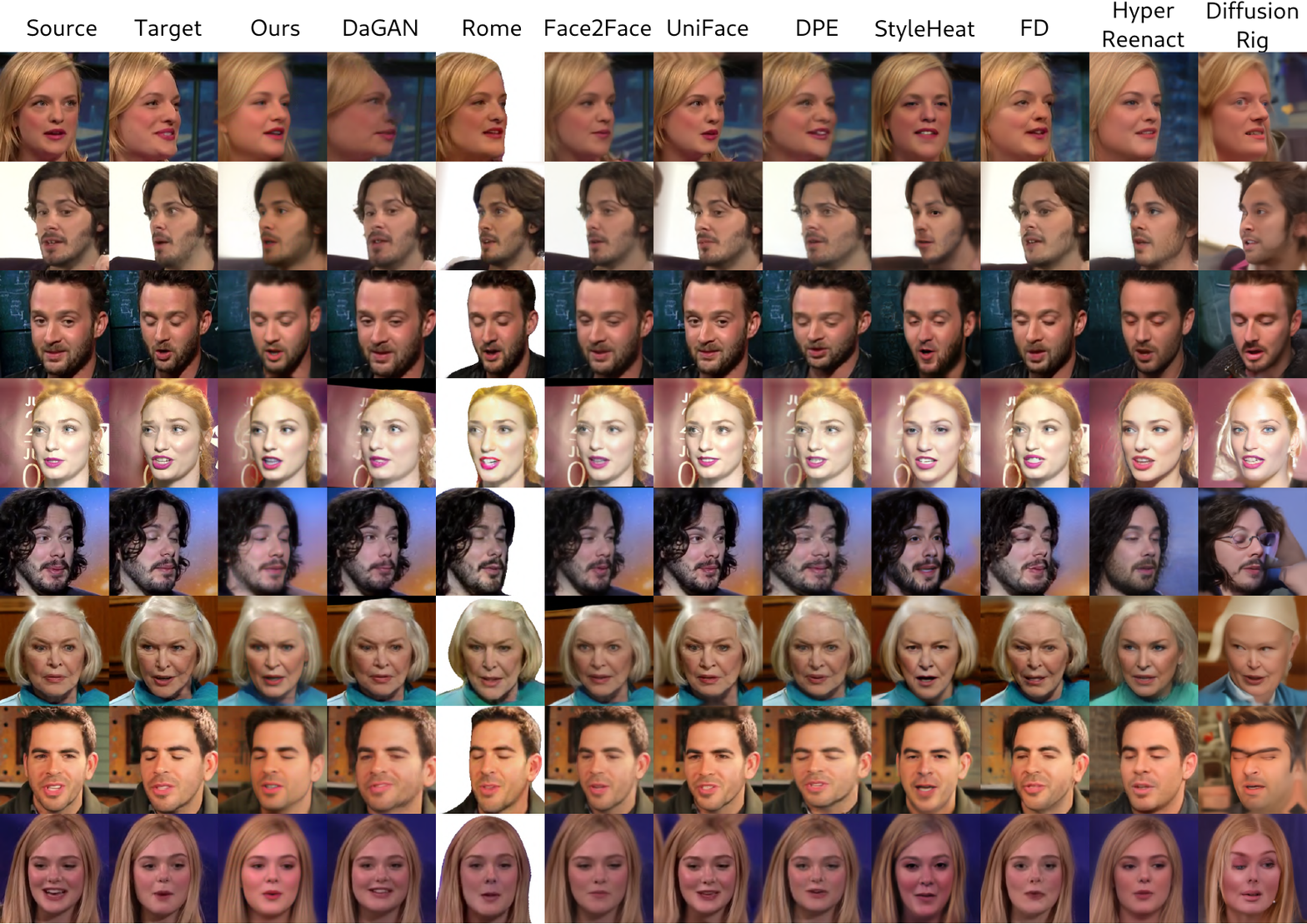}
    \caption{Qualitative comparisons on self reenactment using the VoxCeleb1~\cite{Nagrani17} dataset. The first and second column show the source and target faces respectively.}
    \label{supfig:comparisons_self_vox1}
\end{figure*}

\begin{figure*}[t]
    \centering
    \includegraphics[width=1.0\textwidth]{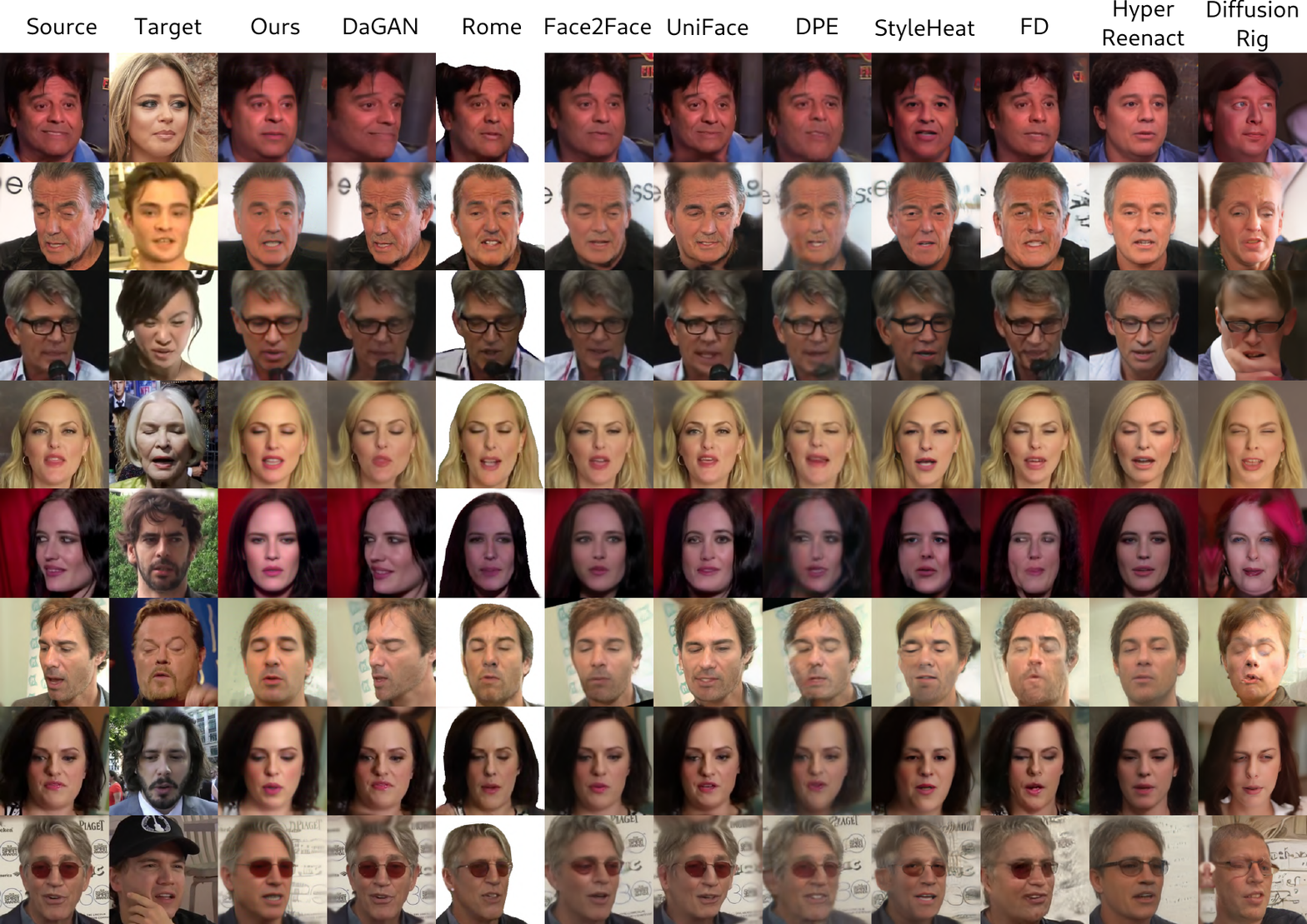}
    \caption{Qualitative comparisons on cross-subject reenactment using the VoxCeleb1~\cite{Nagrani17} dataset. The first and second column show the source and target faces respectively. Our method can preserve the source identity and appearance, while accurately transferring the head pose and the expression of the target faces.}
    \label{supfig:comparisons_cross_vox1}
\end{figure*}

\begin{figure*}[t]
    \centering
    \includegraphics[width=1.0\textwidth]{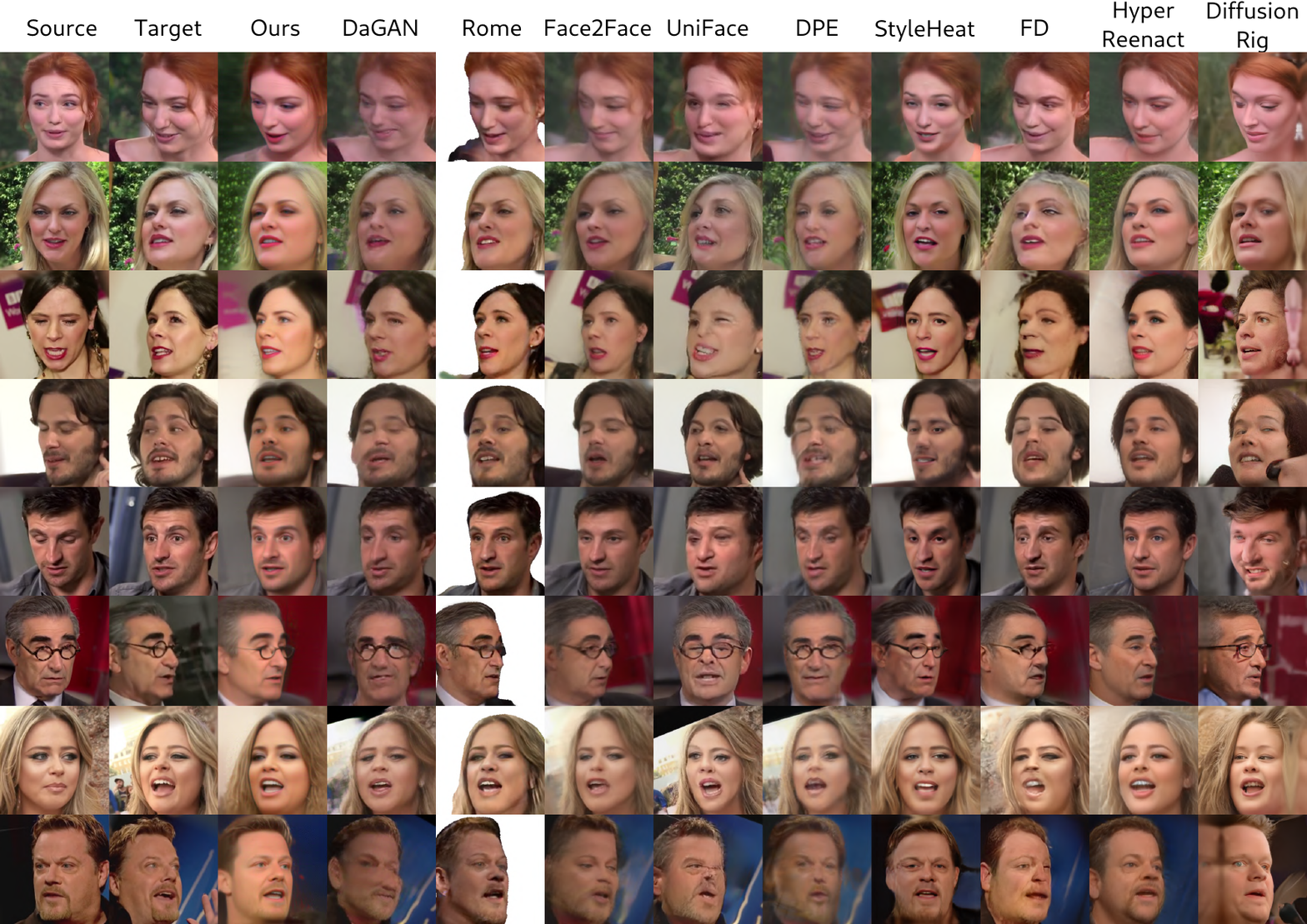}
    \caption{Qualitative comparisons on self reenactment using the benchmark dataset with large head pose differences released by the authors of HyperReenact~\cite{bounareli2023hyperreenact}. The first and second column show the source and target faces respectively. As shown our method even on the challenging task of extreme head pose variations is able to generate realistic and artifact-free images.}
    \label{supfig:comparisons_benchmark}
\end{figure*}

\begin{figure*}[t]
    \centering
    \includegraphics[width=1.0\textwidth]{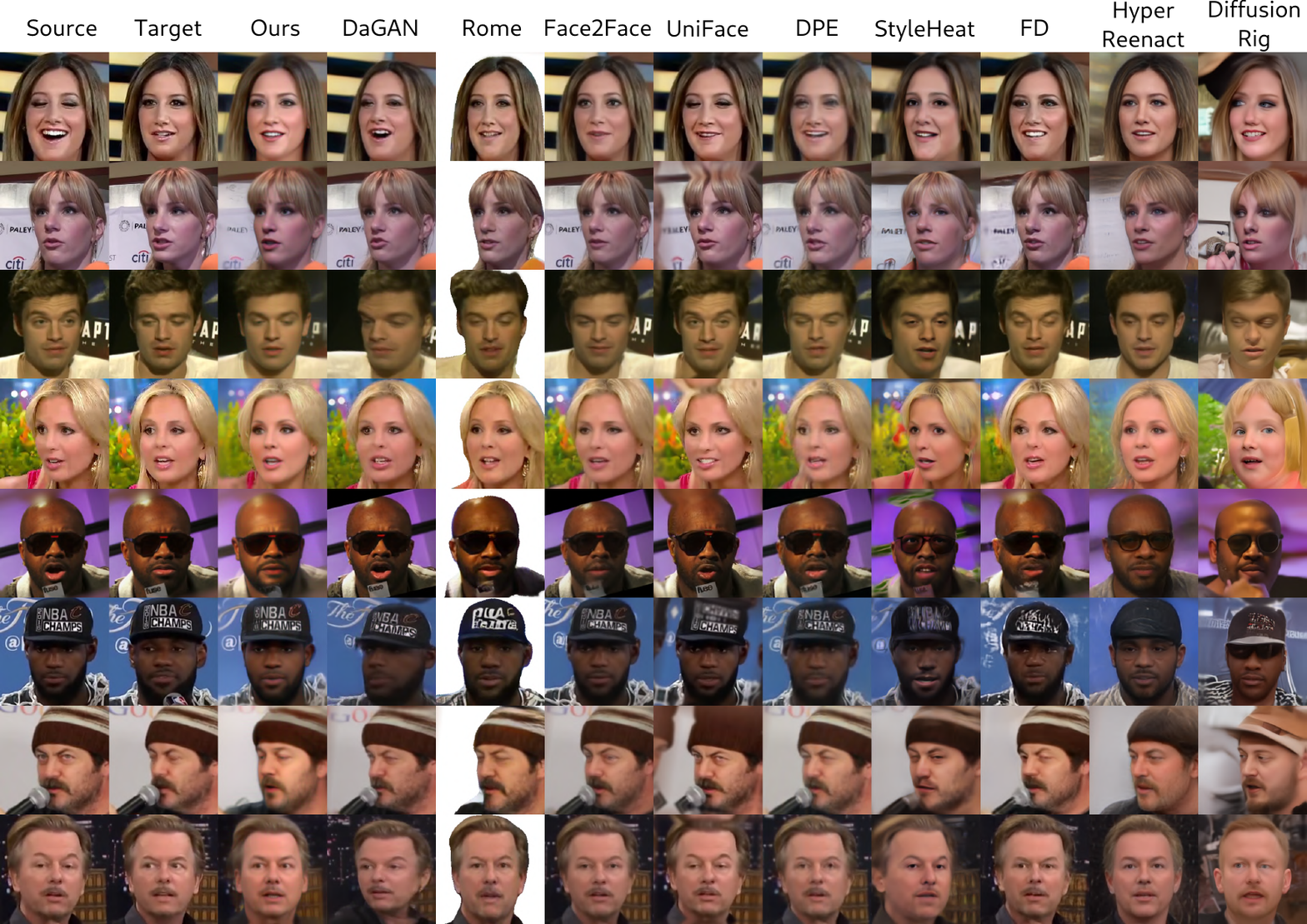}
    \caption{Qualitative comparisons on self reenactment using the VoxCeleb2~\cite{Nagrani17} dataset. The first and second column show the source and target faces respectively. As shown our method can successfully reconstruct the appearance details (e.g., glasses and hats) of the source images.}
    \label{supfig:comparisons_self_vox2}
\end{figure*}

\begin{figure*}[t]
    \centering
    \includegraphics[width=1.0\textwidth]{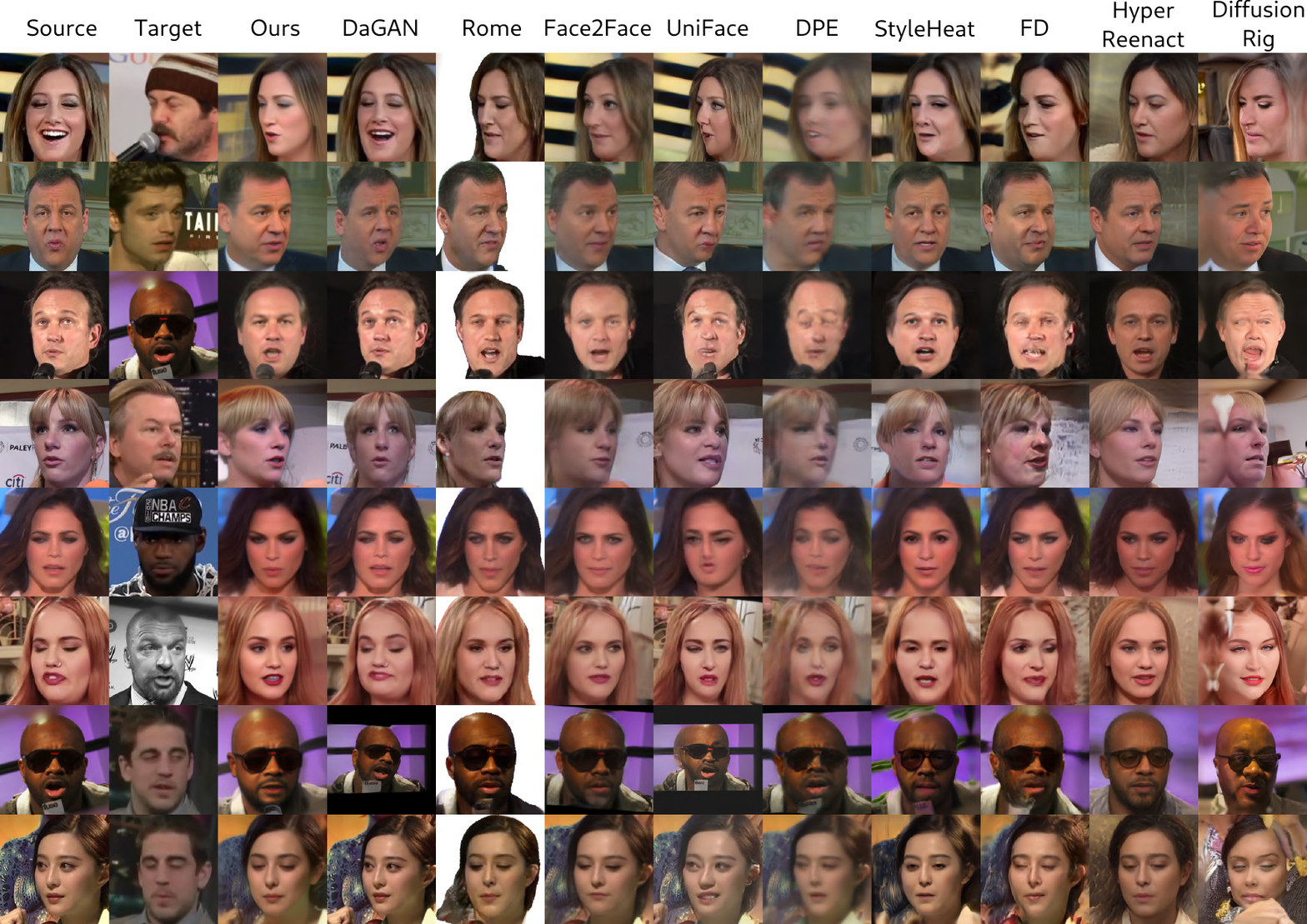}
    \caption{Qualitative comparisons on cross-subject reenactment using the VoxCeleb2~\cite{Nagrani17} dataset. The first and second column show the source and target faces respectively. Compared to the other methods, our approach can generate realistic and artifact-free images.}
    \label{supfig:comparisons_cross_vox2}
\end{figure*}

\begin{figure*}[t]
    \centering
    \includegraphics[width=0.9\textwidth]{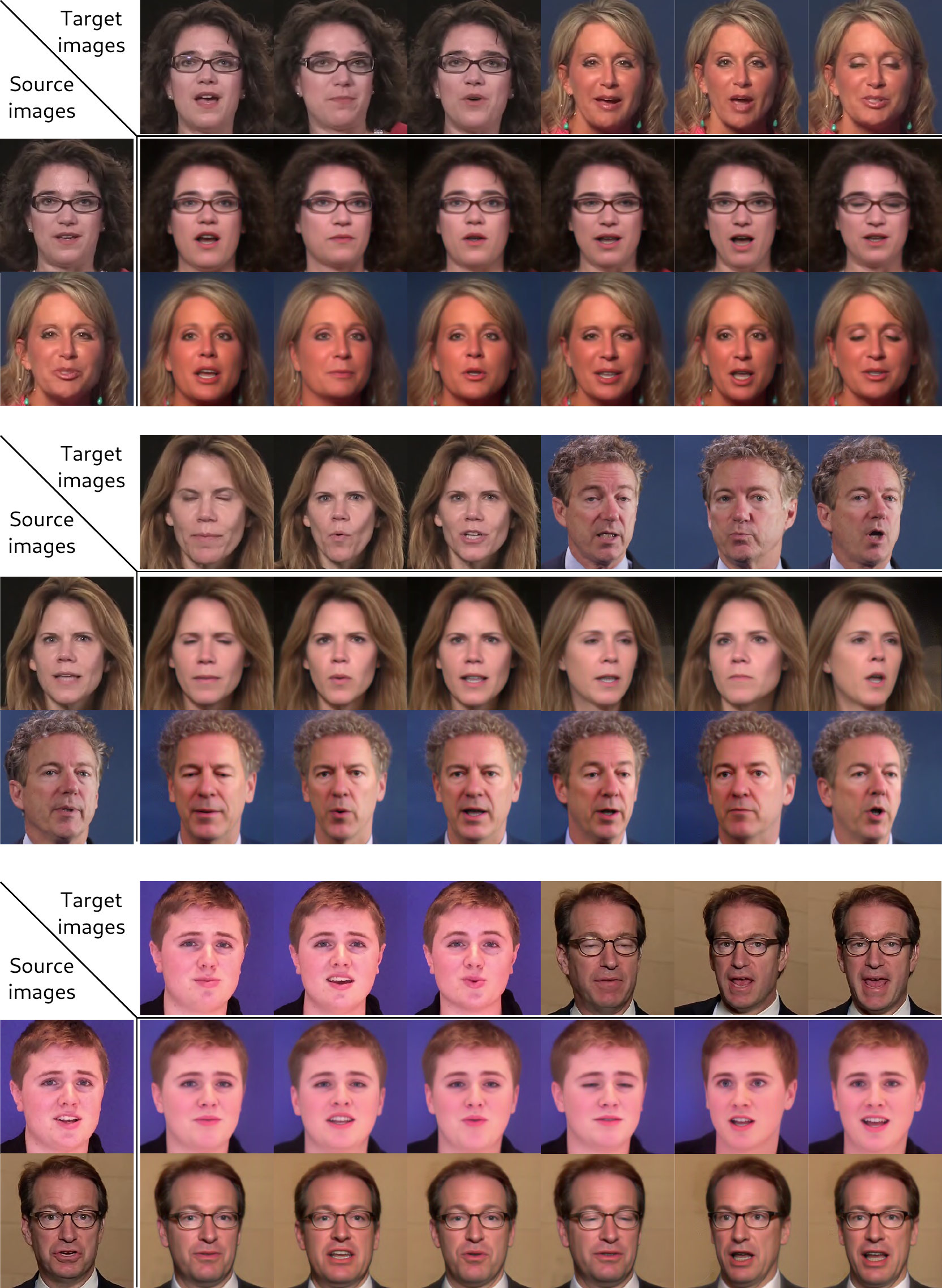}
    \caption{Qualitative comparisons on self and cross-subject reenactment using the HDTF~\cite{zhang2021flow} dataset. Our method generates realistic images that maintain the source identity and appearance both on self and on cross-subject reenactment.}
    \label{suppfig:comparisons_hdtf}
\end{figure*}

\section{Limitations}\label{sec:limitations}

    In this work, we present a method for neural face reenactment based on a pre-trained diffusion model. In our experimental results (Sect.~4.2 of the main paper), we show that the proposed framework is able to accurately transfer the head pose and facial expressions of the target faces, while at the same time the identity and the appearance characteristics of the source face are faithfully preserved. Nevertheless, as shown in Table~\ref{supptable:temporal} in this document, our method, which builds on a DDIM-based method~\cite{Song2021ddim,preechakul2022diffusionauto}, exhibits slower inference time compared to GAN- and StyleGAN2-based methods, as expected. Additionally, we observe that, in some cases, the reenacted images appear brighter than the real ones, as illustrated in Fig.~\ref{supfig:limitations}. Notably, these changes in brightness are particularly prominent in darker, low-resolution videos. We attribute this to the adopted DDIM sampler, which, since it has been pre-trained on the FFHQ dataset~\cite{karras2019style}, differs significantly from the VoxCeleb datasets~\cite{Nagrani17,Chung18b} in terms of image quality.

\section{Ethical considerations}

Neural face reenactment methods can be used in various domains, such as in film and video production, video conferencing, video games, and in augmented and virtual reality applications. However, we recognize that it can also be used for malicious purposes to generate deep fake videos and spread false information, that could potentially harm individuals and the society. Nowadays, there are many neural face reenactment methods that can create high-quality deep fake videos that are indistinguishable from the real ones. As a result, currently there is great interest on deep fake detection methods, that could prevent the malicious use of deep fake generated videos. We will share our model and results with the deepfake detection community to help researchers train more accurate deep fake detection algorithms.  

\begin{figure*}[h]
    \centering
    \includegraphics[width=0.95\textwidth]{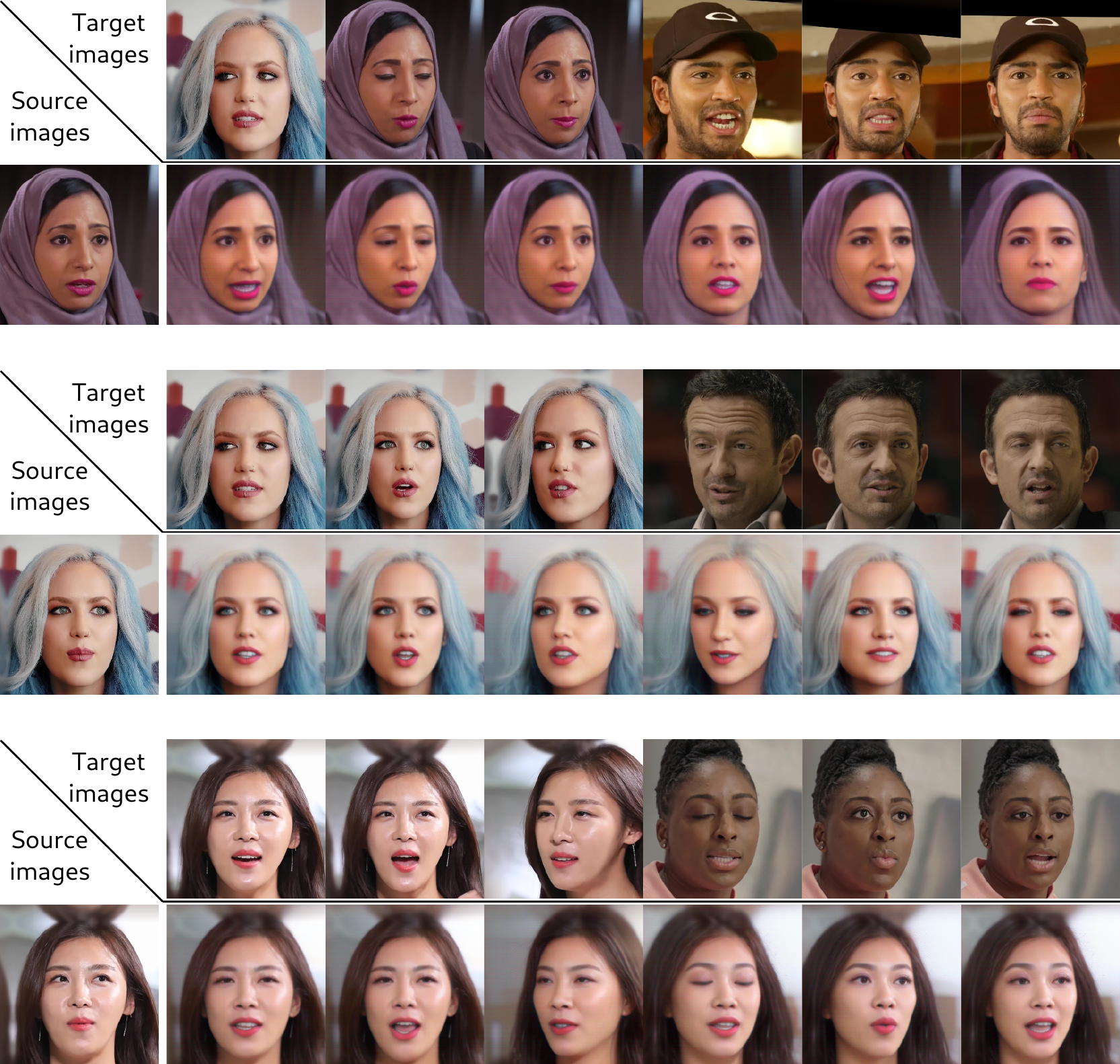}
    \caption{Qualitative comparisons on self and cross-subject reenactment using the VFHQ~\cite{xie2022vfhq} dataset. Our method is able to generalize well on different video datasets and preserve the appearance details of the source images even on the challenging task of cross-subject reenactment.}
    \label{suppfig:comparisons_vfhq}
\end{figure*}

\clearpage
\clearpage
{
    \small
    \bibliographystyle{ieeenat_fullname}
    \bibliography{main}

\begin{thebibliography}{76}
\providecommand{\natexlab}[1]{#1}
\providecommand{\url}[1]{\texttt{#1}}
\expandafter\ifx\csname urlstyle\endcsname\relax
  \providecommand{\doi}[1]{doi: #1}\else
  \providecommand{\doi}{doi: \begingroup \urlstyle{rm}\Url}\fi

\bibitem[Barattin et~al.(2023)Barattin, Tzelepis, Patras, and
  Sebe]{barattin2023attribute}
Simone Barattin, Christos Tzelepis, Ioannis Patras, and Nicu Sebe.
\newblock Attribute-preserving face dataset anonymization via latent code
  optimization.
\newblock In \emph{Proceedings of the IEEE/CVF Conference on Computer Vision
  and Pattern Recognition}, pages 8001--8010, 2023.

\bibitem[Blanz and Vetter(1999)]{blanz1999morphable}
Volker Blanz and Thomas Vetter.
\newblock A morphable model for the synthesis of 3d faces.
\newblock In \emph{Proceedings of the 26th annual conference on Computer
  graphics and interactive techniques}, 1999.

\bibitem[Bounareli et~al.(2022)Bounareli, Argyriou, and
  Tzimiropoulos]{bounareli2022finding}
Stella Bounareli, Vasileios Argyriou, and Georgios Tzimiropoulos.
\newblock Finding directions in gan's latent space for neural face reenactment.
\newblock \emph{British Machine Vision Conference (BMVC)}, 2022.

\bibitem[Bounareli et~al.(2023{\natexlab{a}})Bounareli, Tzelepis, Argyriou,
  Patras, and Tzimiropoulos]{bounareli2023hyperreenact}
Stella Bounareli, Christos Tzelepis, Vasileios Argyriou, Ioannis Patras, and
  Georgios Tzimiropoulos.
\newblock Hyperreenact: one-shot reenactment via jointly learning to refine and
  retarget faces.
\newblock In \emph{Proceedings of the IEEE/CVF International Conference on
  Computer Vision}, pages 7149--7159, 2023{\natexlab{a}}.

\bibitem[Bounareli et~al.(2023{\natexlab{b}})Bounareli, Tzelepis, Argyriou,
  Patras, and Tzimiropoulos]{bounareli2023stylemask}
Stella Bounareli, Christos Tzelepis, Vasileios Argyriou, Ioannis Patras, and
  Georgios Tzimiropoulos.
\newblock Stylemask: Disentangling the style space of stylegan2 for neural face
  reenactment.
\newblock In \emph{2023 IEEE 17th International Conference on Automatic Face
  and Gesture Recognition (FG)}, pages 1--8. IEEE, 2023{\natexlab{b}}.

\bibitem[Bulat and Tzimiropoulos(2017)]{bulat2017far}
Adrian Bulat and Georgios Tzimiropoulos.
\newblock How far are we from solving the 2d \& 3d face alignment problem?(and
  a dataset of 230,000 3d facial landmarks).
\newblock In \emph{Proceedings of the IEEE international conference on computer
  vision}, pages 1021--1030, 2017.

\bibitem[Burkov et~al.(2020)Burkov, Pasechnik, Grigorev, and
  Lempitsky]{burkov2020neural}
Egor Burkov, Igor Pasechnik, Artur Grigorev, and Victor Lempitsky.
\newblock Neural head reenactment with latent pose descriptors.
\newblock In \emph{CVPR}, 2020.

\bibitem[Chen and Lathuili{\`e}re(2023)]{chen2023face}
Xiangyi Chen and St{\'e}phane Lathuili{\`e}re.
\newblock Face aging via diffusion-based editing.
\newblock \emph{arXiv preprint arXiv:2309.11321}, 2023.

\bibitem[Chung et~al.(2018)Chung, Nagrani, and Zisserman]{Chung18b}
J.~S. Chung, A. Nagrani, and A. Zisserman.
\newblock Voxceleb2: Deep speaker recognition.
\newblock In \emph{INTERSPEECH}, 2018.

\bibitem[Dan{\v{e}}{\v{c}}ek et~al.(2022)Dan{\v{e}}{\v{c}}ek, Black, and
  Bolkart]{danvevcek2022emoca}
Radek Dan{\v{e}}{\v{c}}ek, Michael~J Black, and Timo Bolkart.
\newblock Emoca: Emotion driven monocular face capture and animation.
\newblock In \emph{Proceedings of the IEEE/CVF Conference on Computer Vision
  and Pattern Recognition}, pages 20311--20322, 2022.

\bibitem[Deng et~al.(2019)Deng, Guo, Xue, and Zafeiriou]{deng2019arcface}
Jiankang Deng, Jia Guo, Niannan Xue, and Stefanos Zafeiriou.
\newblock Arcface: Additive angular margin loss for deep face recognition.
\newblock In \emph{Proceedings of the IEEE/CVF conference on computer vision
  and pattern recognition}, pages 4690--4699, 2019.

\bibitem[Dhariwal and Nichol(2021)]{dhariwal2021diffusion}
Prafulla Dhariwal and Alexander Nichol.
\newblock Diffusion models beat gans on image synthesis.
\newblock \emph{Advances in neural information processing systems},
  34:\penalty0 8780--8794, 2021.

\bibitem[Ding et~al.(2023)Ding, Zhang, Xia, Jebe, Tu, and
  Zhang]{ding2023diffusionrig}
Zheng Ding, Xuaner Zhang, Zhihao Xia, Lars Jebe, Zhuowen Tu, and Xiuming Zhang.
\newblock Diffusionrig: Learning personalized priors for facial appearance
  editing.
\newblock In \emph{Proceedings of the IEEE/CVF Conference on Computer Vision
  and Pattern Recognition}, pages 12736--12746, 2023.

\bibitem[Doukas et~al.(2020)Doukas, Zafeiriou, and
  Sharmanska]{doukas2020headgan}
Michail~Christos Doukas, Stefanos Zafeiriou, and Viktoriia Sharmanska.
\newblock Headgan: Video-and-audio-driven talking head synthesis.
\newblock \emph{arXiv preprint arXiv:2012.08261}, 2020.

\bibitem[Du et~al.(2023)Du, Chen, He, Tan, Chen, Yu, Zhao, and Bian]{du2023dae}
Chenpeng Du, Qi Chen, Tianyu He, Xu Tan, Xie Chen, Kai Yu, Sheng Zhao, and
  Jiang Bian.
\newblock Dae-talker: High fidelity speech-driven talking face generation with
  diffusion autoencoder.
\newblock In \emph{Proceedings of the 31st ACM International Conference on
  Multimedia}, pages 4281--4289, 2023.

\bibitem[Feng et~al.(2021)Feng, Feng, Black, and Bolkart]{feng2020deca}
Yao Feng, Haiwen Feng, Michael~J Black, and Timo Bolkart.
\newblock Learning an animatable detailed 3d face model from in-the-wild
  images.
\newblock \emph{ACM Transactions on Graphics (TOG)}, 40\penalty0 (4):\penalty0
  1--13, 2021.

\bibitem[Gan et~al.(2023)Gan, Yang, Yue, Sun, and Yang]{gan2023efficient}
Yuan Gan, Zongxin Yang, Xihang Yue, Lingyun Sun, and Yi Yang.
\newblock Efficient emotional adaptation for audio-driven talking-head
  generation.
\newblock In \emph{Proceedings of the IEEE/CVF International Conference on
  Computer Vision}, pages 22634--22645, 2023.

\bibitem[Goodfellow et~al.(2014)Goodfellow, Pouget-Abadie, Mirza, Xu,
  Warde-Farley, Ozair, Courville, and Bengio]{goodfellow2014generative}
Ian Goodfellow, Jean Pouget-Abadie, Mehdi Mirza, Bing Xu, David Warde-Farley,
  Sherjil Ozair, Aaron Courville, and Yoshua Bengio.
\newblock Generative adversarial nets.
\newblock In \emph{Advances in neural information processing systems}, pages
  2672--2680, 2014.

\bibitem[Guan et~al.(2023)Guan, Zhang, Zhou, Hu, Wang, He, Feng, Liu, Ding,
  Liu, et~al.]{guan2023stylesync}
Jiazhi Guan, Zhanwang Zhang, Hang Zhou, Tianshu Hu, Kaisiyuan Wang, Dongliang
  He, Haocheng Feng, Jingtuo Liu, Errui Ding, Ziwei Liu, et~al.
\newblock Stylesync: High-fidelity generalized and personalized lip sync in
  style-based generator.
\newblock In \emph{Proceedings of the IEEE/CVF Conference on Computer Vision
  and Pattern Recognition}, pages 1505--1515, 2023.

\bibitem[Ha et~al.(2017)Ha, Dai, and Le]{hypernets}
David Ha, Andrew~M. Dai, and Quoc~V. Le.
\newblock Hypernetworks.
\newblock In \emph{5th International Conference on Learning Representations,
  {ICLR} 2017, Toulon, France, April 24-26, 2017, Conference Track
  Proceedings}, 2017.

\bibitem[Ho et~al.(2020)Ho, Jain, and Abbeel]{ho2020denoising}
Jonathan Ho, Ajay Jain, and Pieter Abbeel.
\newblock Denoising diffusion probabilistic models.
\newblock \emph{Advances in neural information processing systems},
  33:\penalty0 6840--6851, 2020.

\bibitem[Hong et~al.(2022)Hong, Zhang, Shen, and Xu]{hong2022depth}
Fa-Ting Hong, Longhao Zhang, Li Shen, and Dan Xu.
\newblock Depth-aware generative adversarial network for talking head video
  generation.
\newblock In \emph{Proceedings of the IEEE/CVF conference on computer vision
  and pattern recognition}, pages 3397--3406, 2022.

\bibitem[Hsu et~al.(2022)Hsu, Tsai, and Wu]{hsu2022dual}
Gee-Sern Hsu, Chun-Hung Tsai, and Hung-Yi Wu.
\newblock Dual-generator face reenactment.
\newblock In \emph{Proceedings of the IEEE/CVF Conference on Computer Vision
  and Pattern Recognition}, pages 642--650, 2022.

\bibitem[Huang et~al.(2023)Huang, Chan, Jiang, and Liu]{huang2023collaborative}
Ziqi Huang, Kelvin~CK Chan, Yuming Jiang, and Ziwei Liu.
\newblock Collaborative diffusion for multi-modal face generation and editing.
\newblock In \emph{Proceedings of the IEEE/CVF Conference on Computer Vision
  and Pattern Recognition}, pages 6080--6090, 2023.

\bibitem[Härkönen et~al.(2020)Härkönen, Hertzmann, Lehtinen, and
  Paris]{2020ganspace}
Erik Härkönen, Aaron Hertzmann, Jaakko Lehtinen, and Sylvain Paris.
\newblock Ganspace: Discovering interpretable gan controls.
\newblock In \emph{Proc. NeurIPS}, 2020.

\bibitem[Johnson et~al.(2016)Johnson, Alahi, and
  Fei-Fei]{johnson2016perceptual}
Justin Johnson, Alexandre Alahi, and Li Fei-Fei.
\newblock Perceptual losses for real-time style transfer and super-resolution.
\newblock In \emph{Computer Vision--ECCV 2016: 14th European Conference,
  Amsterdam, The Netherlands, October 11-14, 2016, Proceedings, Part II 14},
  pages 694--711. Springer, 2016.

\bibitem[Karras et~al.(2018)Karras, Aila, Laine, and
  Lehtinen]{karras2017progressive}
Tero Karras, Timo Aila, Samuli Laine, and Jaakko Lehtinen.
\newblock Progressive growing of gans for improved quality, stability, and
  variation.
\newblock In \emph{6th International Conference on Learning Representations,
  {ICLR} 2018, Vancouver, BC, Canada, April 30 - May 3, 2018, Conference Track
  Proceedings}, 2018.

\bibitem[Karras et~al.(2019)Karras, Laine, and Aila]{karras2019style}
Tero Karras, Samuli Laine, and Timo Aila.
\newblock A style-based generator architecture for generative adversarial
  networks.
\newblock In \emph{Proceedings of the IEEE/CVF Conference on Computer Vision
  and Pattern Recognition}, pages 4401--4410, 2019.

\bibitem[Karras et~al.(2020)Karras, Laine, Aittala, Hellsten, Lehtinen, and
  Aila]{karras2020analyzing}
Tero Karras, Samuli Laine, Miika Aittala, Janne Hellsten, Jaakko Lehtinen, and
  Timo Aila.
\newblock Analyzing and improving the image quality of stylegan.
\newblock In \emph{Proceedings of the IEEE/CVF Conference on Computer Vision
  and Pattern Recognition}, pages 8110--8119, 2020.

\bibitem[Khakhulin et~al.(2022)Khakhulin, Sklyarova, Lempitsky, and
  Zakharov]{khakhulin2022rome}
Taras Khakhulin, Vanessa Sklyarova, Victor Lempitsky, and Egor Zakharov.
\newblock Realistic one-shot mesh-based head avatars.
\newblock In \emph{European Conference on Computer Vision}, pages 345--362.
  Springer, 2022.

\bibitem[Kim et~al.(2022)Kim, Kwon, and Ye]{kim2022diffusionclip}
Gwanghyun Kim, Taesung Kwon, and Jong~Chul Ye.
\newblock Diffusionclip: Text-guided diffusion models for robust image
  manipulation.
\newblock In \emph{Proceedings of the IEEE/CVF Conference on Computer Vision
  and Pattern Recognition}, pages 2426--2435, 2022.

\bibitem[Kim et~al.(2023)Kim, Shim, Kim, Choi, Kim, and Yang]{kim2023diffusion}
Gyeongman Kim, Hajin Shim, Hyunsu Kim, Yunjey Choi, Junho Kim, and Eunho Yang.
\newblock Diffusion video autoencoders: Toward temporally consistent face video
  editing via disentangled video encoding.
\newblock In \emph{Proceedings of the IEEE/CVF Conference on Computer Vision
  and Pattern Recognition}, pages 6091--6100, 2023.

\bibitem[Loshchilov and Hutter(2019)]{adamw}
Ilya Loshchilov and Frank Hutter.
\newblock Decoupled weight decay regularization.
\newblock In \emph{7th International Conference on Learning Representations,
  {ICLR} 2019, New Orleans, LA, USA, May 6-9, 2019}. OpenReview.net, 2019.

\bibitem[Lugmayr et~al.(2022)Lugmayr, Danelljan, Romero, Yu, Timofte, and
  Van~Gool]{lugmayr2022repaint}
Andreas Lugmayr, Martin Danelljan, Andres Romero, Fisher Yu, Radu Timofte, and
  Luc Van~Gool.
\newblock Repaint: Inpainting using denoising diffusion probabilistic models.
\newblock In \emph{Proceedings of the IEEE/CVF Conference on Computer Vision
  and Pattern Recognition}, pages 11461--11471, 2022.

\bibitem[Meng et~al.(2022)Meng, He, Song, Song, Wu, Zhu, and Ermon]{sedit2022}
Chenlin Meng, Yutong He, Yang Song, Jiaming Song, Jiajun Wu, Jun{-}Yan Zhu, and
  Stefano Ermon.
\newblock Sdedit: Guided image synthesis and editing with stochastic
  differential equations.
\newblock In \emph{The Tenth International Conference on Learning
  Representations, {ICLR} 2022, Virtual Event, April 25-29, 2022}.
  OpenReview.net, 2022.

\bibitem[Meshry et~al.(2021)Meshry, Suri, Davis, and
  Shrivastava]{meshry2021learned}
Moustafa Meshry, Saksham Suri, Larry~S Davis, and Abhinav Shrivastava.
\newblock Learned spatial representations for few-shot talking-head synthesis.
\newblock In \emph{Proceedings of the IEEE/CVF International Conference on
  Computer Vision}, pages 13829--13838, 2021.

\bibitem[Mokady et~al.(2023)Mokady, Hertz, Aberman, Pritch, and
  Cohen-Or]{mokady2023null}
Ron Mokady, Amir Hertz, Kfir Aberman, Yael Pritch, and Daniel Cohen-Or.
\newblock Null-text inversion for editing real images using guided diffusion
  models.
\newblock In \emph{Proceedings of the IEEE/CVF Conference on Computer Vision
  and Pattern Recognition}, pages 6038--6047, 2023.

\bibitem[Nagrani et~al.(2017)Nagrani, Chung, and Zisserman]{Nagrani17}
A. Nagrani, J.~S. Chung, and A. Zisserman.
\newblock Voxceleb: a large-scale speaker identification dataset.
\newblock In \emph{INTERSPEECH}, 2017.

\bibitem[Oldfield et~al.(2023)Oldfield, Tzelepis, Panagakis, Nicolaou, and
  Patras]{oldfield2023panda}
James Oldfield, Christos Tzelepis, Yannis Panagakis, Mihalis Nicolaou, and
  Ioannis Patras.
\newblock Panda: Unsupervised learning of parts and appearances in the feature
  maps of gans.
\newblock 2023.

\bibitem[Oorloff and Yacoob(2023)]{oorloff2023robust}
Trevine Oorloff and Yaser Yacoob.
\newblock Robust one-shot face video re-enactment using hybrid latent spaces of
  stylegan2.
\newblock In \emph{Proceedings of the IEEE/CVF International Conference on
  Computer Vision}, pages 20947--20957, 2023.

\bibitem[Pan et~al.(2023)Pan, Gherardi, Xie, and Huang]{pan2023effective}
Zhihong Pan, Riccardo Gherardi, Xiufeng Xie, and Stephen Huang.
\newblock Effective real image editing with accelerated iterative diffusion
  inversion.
\newblock In \emph{Proceedings of the IEEE/CVF International Conference on
  Computer Vision}, pages 15912--15921, 2023.

\bibitem[Pang et~al.(2023)Pang, Zhang, Quan, Fan, Cun, Shan, and
  Yan]{pang2023dpe}
Youxin Pang, Yong Zhang, Weize Quan, Yanbo Fan, Xiaodong Cun, Ying Shan, and
  Dong-ming Yan.
\newblock Dpe: Disentanglement of pose and expression for general video
  portrait editing.
\newblock In \emph{Proceedings of the IEEE/CVF Conference on Computer Vision
  and Pattern Recognition}, pages 427--436, 2023.

\bibitem[Patashnik et~al.(2021)Patashnik, Wu, Shechtman, Cohen-Or, and
  Lischinski]{patashnik2021styleclip}
Or Patashnik, Zongze Wu, Eli Shechtman, Daniel Cohen-Or, and Dani Lischinski.
\newblock Styleclip: Text-driven manipulation of stylegan imagery.
\newblock In \emph{Proceedings of the IEEE/CVF International Conference on
  Computer Vision}, 2021.

\bibitem[Ponglertnapakorn et~al.(2023)Ponglertnapakorn, Tritrong, and
  Suwajanakorn]{ponglertnapakorn2023difareli}
Puntawat Ponglertnapakorn, Nontawat Tritrong, and Supasorn Suwajanakorn.
\newblock Difareli: Diffusion face relighting.
\newblock \emph{arXiv preprint arXiv:2304.09479}, 2023.

\bibitem[Prajwal et~al.(2020)Prajwal, Mukhopadhyay, Namboodiri, and
  Jawahar]{prajwal2020lip}
KR Prajwal, Rudrabha Mukhopadhyay, Vinay~P Namboodiri, and CV Jawahar.
\newblock A lip sync expert is all you need for speech to lip generation in the
  wild.
\newblock In \emph{Proceedings of the 28th ACM international conference on
  multimedia}, pages 484--492, 2020.

\bibitem[Preechakul et~al.(2022)Preechakul, Chatthee, Wizadwongsa, and
  Suwajanakorn]{preechakul2022diffusionauto}
Konpat Preechakul, Nattanat Chatthee, Suttisak Wizadwongsa, and Supasorn
  Suwajanakorn.
\newblock Diffusion autoencoders: Toward a meaningful and decodable
  representation.
\newblock In \emph{Proceedings of the IEEE/CVF Conference on Computer Vision
  and Pattern Recognition}, pages 10619--10629, 2022.

\bibitem[Ren et~al.(2021)Ren, Li, Chen, Li, and Liu]{ren2021pirenderer}
Yurui Ren, Ge Li, Yuanqi Chen, Thomas~H Li, and Shan Liu.
\newblock Pirenderer: Controllable portrait image generation via semantic
  neural rendering.
\newblock In \emph{Proceedings of the IEEE/CVF International Conference on
  Computer Vision}, pages 13759--13768, 2021.

\bibitem[Roich et~al.(2021)Roich, Mokady, Bermano, and
  Cohen-Or]{roich2021pivotal}
Daniel Roich, Ron Mokady, Amit~H Bermano, and Daniel Cohen-Or.
\newblock Pivotal tuning for latent-based editing of real images.
\newblock \emph{arXiv preprint arXiv:2106.05744}, 2021.

\bibitem[Rombach et~al.(2022)Rombach, Blattmann, Lorenz, Esser, and
  Ommer]{rombach2022high}
Robin Rombach, Andreas Blattmann, Dominik Lorenz, Patrick Esser, and Bj{\"o}rn
  Ommer.
\newblock High-resolution image synthesis with latent diffusion models.
\newblock In \emph{Proceedings of the IEEE/CVF conference on computer vision
  and pattern recognition}, pages 10684--10695, 2022.

\bibitem[Ronneberger et~al.(2015)Ronneberger, Fischer, and
  Brox]{ronneberger2015u}
Olaf Ronneberger, Philipp Fischer, and Thomas Brox.
\newblock U-net: Convolutional networks for biomedical image segmentation.
\newblock In \emph{Medical Image Computing and Computer-Assisted
  Intervention--MICCAI 2015: 18th International Conference, Munich, Germany,
  October 5-9, 2015, Proceedings, Part III 18}, pages 234--241. Springer, 2015.

\bibitem[Shen et~al.(2023)Shen, Zhao, Meng, Li, Zhu, Zhou, and
  Lu]{shen2023difftalk}
Shuai Shen, Wenliang Zhao, Zibin Meng, Wanhua Li, Zheng Zhu, Jie Zhou, and
  Jiwen Lu.
\newblock Difftalk: Crafting diffusion models for generalized audio-driven
  portraits animation.
\newblock In \emph{Proceedings of the IEEE/CVF Conference on Computer Vision
  and Pattern Recognition}, pages 1982--1991, 2023.

\bibitem[Siarohin et~al.(2019)Siarohin, Lathuili{\`e}re, Tulyakov, Ricci, and
  Sebe]{siarohin2019first}
Aliaksandr Siarohin, St{\'e}phane Lathuili{\`e}re, Sergey Tulyakov, Elisa
  Ricci, and Nicu Sebe.
\newblock First order motion model for image animation.
\newblock \emph{Advances in Neural Information Processing Systems},
  32:\penalty0 7137--7147, 2019.

\bibitem[Song et~al.(2021)Song, Meng, and Ermon]{Song2021ddim}
Jiaming Song, Chenlin Meng, and Stefano Ermon.
\newblock Denoising diffusion implicit models.
\newblock In \emph{9th International Conference on Learning Representations,
  {ICLR} 2021, Virtual Event, Austria, May 3-7, 2021}, 2021.

\bibitem[Stypu{\l}kowski et~al.(2023)Stypu{\l}kowski, Vougioukas, He,
  Zi{\k{e}}ba, Petridis, and Pantic]{stypulkowski2023diffused}
Micha{\l} Stypu{\l}kowski, Konstantinos Vougioukas, Sen He, Maciej Zi{\k{e}}ba,
  Stavros Petridis, and Maja Pantic.
\newblock Diffused heads: Diffusion models beat gans on talking-face
  generation.
\newblock \emph{arXiv preprint arXiv:2301.03396}, 2023.

\bibitem[Tov et~al.(2021)Tov, Alaluf, Nitzan, Patashnik, and
  Cohen-Or]{tov2021designing}
Omer Tov, Yuval Alaluf, Yotam Nitzan, Or Patashnik, and Daniel Cohen-Or.
\newblock Designing an encoder for stylegan image manipulation.
\newblock \emph{ACM Transactions on Graphics (TOG)}, 40\penalty0 (4):\penalty0
  1--14, 2021.

\bibitem[Tzaban et~al.(2022)Tzaban, Mokady, Gal, Bermano, and
  Cohen-Or]{tzaban2022stitch}
Rotem Tzaban, Ron Mokady, Rinon Gal, Amit Bermano, and Daniel Cohen-Or.
\newblock Stitch it in time: Gan-based facial editing of real videos.
\newblock In \emph{SIGGRAPH Asia 2022 Conference Papers}, pages 1--9, 2022.

\bibitem[Tzelepis et~al.(2021)Tzelepis, Tzimiropoulos, and
  Patras]{tzelepis2021warpedganspace}
Christos Tzelepis, Georgios Tzimiropoulos, and Ioannis Patras.
\newblock Warpedganspace: Finding non-linear rbf paths in gan latent space.
\newblock In \emph{Proceedings of the IEEE/CVF International Conference on
  Computer Vision}, pages 6393--6402, 2021.

\bibitem[Tzelepis et~al.(2022)Tzelepis, Oldfield, Tzimiropoulos, and
  Patras]{tzelepis2022contraclip}
Christos Tzelepis, James Oldfield, Georgios Tzimiropoulos, and Ioannis Patras.
\newblock Contraclip: Interpretable gan generation driven by pairs of
  contrasting sentences.
\newblock \emph{arXiv preprint arXiv:2206.02104}, 2022.

\bibitem[Voynov and Babenko(2020)]{voynov2020unsupervised}
Andrey Voynov and Artem Babenko.
\newblock Unsupervised discovery of interpretable directions in the gan latent
  space.
\newblock In \emph{International Conference on Machine Learning}, pages
  9786--9796. PMLR, 2020.

\bibitem[Wang et~al.(2024)Wang, Bai, Wang, Qin, and Chen]{wang2024instantid}
Qixun Wang, Xu Bai, Haofan Wang, Zekui Qin, and Anthony Chen.
\newblock Instantid: Zero-shot identity-preserving generation in seconds.
\newblock \emph{arXiv preprint arXiv:2401.07519}, 2024.

\bibitem[Wang et~al.(2021)Wang, Mallya, and Liu]{wang2021one}
Ting-Chun Wang, Arun Mallya, and Ming-Yu Liu.
\newblock One-shot free-view neural talking-head synthesis for video
  conferencing.
\newblock In \emph{Proceedings of the IEEE/CVF Conference on Computer Vision
  and Pattern Recognition}, pages 10039--10049, 2021.

\bibitem[Xie et~al.(2022)Xie, Wang, Zhang, Dong, and Shan]{xie2022vfhq}
Liangbin Xie, Xintao Wang, Honglun Zhang, Chao Dong, and Ying Shan.
\newblock Vfhq: A high-quality dataset and benchmark for video face
  super-resolution.
\newblock In \emph{Proceedings of the IEEE/CVF Conference on Computer Vision
  and Pattern Recognition}, pages 657--666, 2022.

\bibitem[Xu et~al.(2022)Xu, Zhang, Han, Tian, Zeng, Tai, Wang, Wang, and
  Liu]{xu2022designing}
Chao Xu, Jiangning Zhang, Yue Han, Guanzhong Tian, Xianfang Zeng, Ying Tai,
  Yabiao Wang, Chengjie Wang, and Yong Liu.
\newblock Designing one unified framework for high-fidelity face reenactment
  and swapping.
\newblock In \emph{European Conference on Computer Vision}, pages 54--71.
  Springer, 2022.

\bibitem[Xu et~al.(2023)Xu, Zhu, Zhu, Huang, Zhang, Tai, and
  Liu]{xu2023multimodal}
Chao Xu, Shaoting Zhu, Junwei Zhu, Tianxin Huang, Jiangning Zhang, Ying Tai,
  and Yong Liu.
\newblock Multimodal-driven talking face generation via a unified
  diffusion-based generator.
\newblock \emph{CoRR (2023)}, pages 1--14, 2023.

\bibitem[Yang et~al.(2022)Yang, Chen, Guo, Zhang, Guo, and
  Zhang]{yang2022face2face}
Kewei Yang, Kang Chen, Daoliang Guo, Song-Hai Zhang, Yuan-Chen Guo, and Weidong
  Zhang.
\newblock Face2face $\rho$: Real-time high-resolution one-shot face
  reenactment.
\newblock In \emph{European conference on computer vision}, pages 55--71.
  Springer, 2022.

\bibitem[Ye et~al.(2023)Ye, Zhang, Liu, Han, and Yang]{ye2023ip}
Hu Ye, Jun Zhang, Sibo Liu, Xiao Han, and Wei Yang.
\newblock Ip-adapter: Text compatible image prompt adapter for text-to-image
  diffusion models.
\newblock \emph{arXiv preprint arXiv:2308.06721}, 2023.

\bibitem[Yin et~al.(2022)Yin, Zhang, Cun, Cao, Fan, Wang, Bai, Wu, Wang, and
  Yang]{yin2022styleheat}
Fei Yin, Yong Zhang, Xiaodong Cun, Mingdeng Cao, Yanbo Fan, Xuan Wang, Qingyan
  Bai, Baoyuan Wu, Jue Wang, and Yujiu Yang.
\newblock Styleheat: One-shot high-resolution editable talking face generation
  via pretrained stylegan.
\newblock \emph{arXiv preprint arXiv:2203.04036}, 2022.

\bibitem[Yu et~al.(2021)Yu, Gao, Wang, Yu, Shen, and Sang]{yu2021bisenet}
Changqian Yu, Changxin Gao, Jingbo Wang, Gang Yu, Chunhua Shen, and Nong Sang.
\newblock Bisenet v2: Bilateral network with guided aggregation for real-time
  semantic segmentation.
\newblock \emph{International Journal of Computer Vision}, 129:\penalty0
  3051--3068, 2021.

\bibitem[Yue et~al.(2023)Yue, Guo, Ning, Cui, Zhu, and Yuan]{yue2023chatface}
Dongxu Yue, Qin Guo, Munan Ning, Jiaxi Cui, Yuesheng Zhu, and Li Yuan.
\newblock Chatface: Chat-guided real face editing via diffusion latent space
  manipulation.
\newblock \emph{arXiv preprint arXiv:2305.14742}, 2023.

\bibitem[Zakharov et~al.(2019)Zakharov, Shysheya, Burkov, and
  Lempitsky]{zakharov2019few}
Egor Zakharov, Aliaksandra Shysheya, Egor Burkov, and Victor Lempitsky.
\newblock Few-shot adversarial learning of realistic neural talking head
  models.
\newblock In \emph{Proceedings of the IEEE/CVF International Conference on
  Computer Vision}, pages 9459--9468, 2019.

\bibitem[Zakharov et~al.(2020)Zakharov, Ivakhnenko, Shysheya, and
  Lempitsky]{zakharov2020fast}
Egor Zakharov, Aleksei Ivakhnenko, Aliaksandra Shysheya, and Victor Lempitsky.
\newblock Fast bi-layer neural synthesis of one-shot realistic head avatars.
\newblock In \emph{ECCV}, 2020.

\bibitem[Zhang et~al.(2023)Zhang, Rao, and Agrawala]{zhang2023controlnet}
Lvmin Zhang, Anyi Rao, and Maneesh Agrawala.
\newblock Adding conditional control to text-to-image diffusion models.
\newblock In \emph{Proceedings of the IEEE/CVF International Conference on
  Computer Vision}, pages 3836--3847, 2023.

\bibitem[Zhang et~al.(2020)Zhang, Park, Beeler, Bradley, Tang, and
  Hilliges]{zhang2020eth}
Xucong Zhang, Seonwook Park, Thabo Beeler, Derek Bradley, Siyu Tang, and Otmar
  Hilliges.
\newblock Eth-xgaze: A large scale dataset for gaze estimation under extreme
  head pose and gaze variation.
\newblock In \emph{Computer Vision--ECCV 2020: 16th European Conference,
  Glasgow, UK, August 23--28, 2020, Proceedings, Part V 16}, pages 365--381.
  Springer, 2020.

\bibitem[Zhang et~al.(2021)Zhang, Li, Ding, and Fan]{zhang2021flow}
Zhimeng Zhang, Lincheng Li, Yu Ding, and Changjie Fan.
\newblock Flow-guided one-shot talking face generation with a high-resolution
  audio-visual dataset.
\newblock In \emph{Proceedings of the IEEE/CVF Conference on Computer Vision
  and Pattern Recognition}, pages 3661--3670, 2021.

\bibitem[Zhao and Zhang(2022)]{zhao2022thin}
Jian Zhao and Hui Zhang.
\newblock Thin-plate spline motion model for image animation.
\newblock In \emph{Proceedings of the IEEE/CVF Conference on Computer Vision
  and Pattern Recognition}, pages 3657--3666, 2022.

\bibitem[Zheng et~al.(2022)Zheng, Yang, Zhang, Bao, Chen, Huang, Yuan, Chen,
  Zeng, and Wen]{zheng2022general}
Yinglin Zheng, Hao Yang, Ting Zhang, Jianmin Bao, Dongdong Chen, Yangyu Huang,
  Lu Yuan, Dong Chen, Ming Zeng, and Fang Wen.
\newblock General facial representation learning in a visual-linguistic manner.
\newblock In \emph{Proceedings of the IEEE/CVF Conference on Computer Vision
  and Pattern Recognition}, pages 18697--18709, 2022.

\end{thebibliography}
}


\end{document}